\documentclass{IEEEtran}
\usepackage{cite}
\usepackage[pdftex]{graphicx}
\usepackage{amsmath}
\usepackage{algorithmic}
\usepackage{array}
\usepackage[caption=false,font=footnotesize,labelfont=sf,textfont=sf]{subfig}
\usepackage{fixltx2e}
\usepackage{stfloats}
\usepackage{url}

\usepackage{enumitem}
\usepackage{amssymb}
\usepackage{comment}
\usepackage{color}

\newcolumntype{L}[1]{>{\raggedright\let\newline\\\arraybackslash\hspace{0pt}}m{#1}}
\newcolumntype{C}[1]{>{\centering\let\newline\\\arraybackslash\hspace{0pt}}m{#1}}
\newcommand{\tabincell}[2]{\begin{tabular}{@{}#1@{}}#2\end{tabular}}
\newcommand{\eg}{e.g.}
\newcommand{\ie}{i.e.}

\newcommand{\etal}{et al. }

\begin{document}
\title{Deep Likelihood Network for Image Restoration with Multiple Degradation Levels}

\author{Yiwen Guo, Ming Lu, Wangmeng Zuo, Changshui Zhang,~\IEEEmembership{Fellow,~IEEE}, and Yurong Chen
\thanks{Y. Guo is with ByteDance AI Lab. E-mail: guoyiwen.ai@bytedance.com.}
\thanks{W. Zuo is with the School of Computer Science and Technology, Harbin Institute of Technology, Harbin 150001, China. E-mail: cswmzuo@gmail.com.}
\thanks{M. Lu and Y. Chen are with Intel Labs China, Beijing 100190, China. E-mail: ming1.lu@intel.com, yurong.chen@intel.com.}
\thanks{C. Zhang is with the Institute for Artificial Intelligence, Tsinghua University (THUAI), Beijing National Research Center for Information Science and Technology (BNRist), the Department of Automation, Tsinghua University, Beijing 100084, China. E-mail: zcs@mail.tsinghua.edu.cn}
\thanks{Manuscript received Nov. 12, 2019.}}

\markboth{SUBMISSION TO IEEE TRANSACTIONS ON IMAGE PROCESSING}{}


\maketitle

 \begin{abstract}
Convolutional neural networks have been proven effective in a variety of image restoration tasks. Most state-of-the-art solutions, however, are trained using images with a single particular degradation level, and their performance deteriorates drastically when applied to other degradation settings. In this paper, we propose deep likelihood network (DL-Net), aiming at generalizing off-the-shelf image restoration networks to succeed over a spectrum of degradation levels. We slightly modify an off-the-shelf network by appending a simple recursive module, which is derived from a fidelity term, for disentangling the computation for multiple degradation levels. Extensive experimental results on image inpainting, interpolation, and super-resolution show the effectiveness of our DL-Net.
 \end{abstract}


\IEEEpeerreviewmaketitle

\section{Introduction}\label{sec:int}

\IEEEPARstart{O}{ver} the past few years, deep convolutional neural networks (CNNs) have advanced the state-of-the-art of a variety of image restoration tasks including single image super-resolution (SISR)~\cite{Dong2014}, inpainting~\cite{Pathak2016}, denoising~\cite{Zhang2017}, colorization~\cite{Zhang2016}, etc.
Most state-of-the-art solutions were trained with pairs of manually generated input images and their anticipated restoration outcomes, based on implicit assumptions about the degeneration process.  
Image degradations might be restricted to a presumed level throughout datasets, \eg, a pre-defined shape, size, or location for inpainting regions~\cite{Kohler2014, Yang2017}, and a designated downsampling strategy from high-resolution images~\cite{Dong2014, Kim2016, Tai2017}. 
Such specifications of the input domain entail severe over-fitting in the obtained CNN models~\cite{Gao2017, Zhang2018}.  
That is, they can succeed when the assumptions are fulfilled and the test degradations are limited to such a particular level, but their performance is unassured in practical applications in which multiple degradations exist and more flexible restorations are required. 

A few endeavors have been made to address this problem.
One straightforward solution is to jointly learn restoring images in various settings of degradations. 
Such a na\"{\i}ve method mitigates the problem only to a certain extent.
It may still fail if the degradations vary within a substantial range of difficulties, as demonstrated in the literature~\cite{Gao2017, Zhang2018}. 
Learning to restore multiple levels of image degradations can be regarded as multiple-task, and thus multi-task learning methods can be exploited.
Besides multi-task learning, curriculum learning~\cite{Bengio2009} and self-paced learning~\cite{Jiang2014} can also be adopted to mitigate the issue by controlling the sampling strategies~\cite{Gao2017}, since deep models are believed to learn much better if the training samples corresponding to tasks of different difficulties are allocated in a more reasonable way. 

In order to address the issue, there is also recent work that introduces customized CNN architectures~\cite{Zhang2018, Liu2018, nazeri2019edgeconnect}, with for example modified convolution operations~\cite{Liu2018} or extra generator for edge hallucination~\cite{nazeri2019edgeconnect} in order to obtain better image inpainting results.
These architectures are specifically designed and probably limited to certain tasks. 

In this paper, we aim at directly generalizing off-the-shelf image restoration CNNs, which might be originally presented to restore images under only one particular degradation level, to succeed over a spectrum of degradation levels.
Meanwhile, we also expect to reuse with the main objective functions and network architectures of these models.
In order to achieve this goal, we propose deep likelihood network (DL-Net): a novel method which disentangles\footnote{With our method, input images with different settings of degradations are processed distinctively due to their different degradation kernel $W_i$s and other hyper-parameters, which is what we meant by ``disentangle". } the effect of possible image degradations and enforces high overall likelihood.
The computation procedure of the degradation was cast into a recursive module and can be readily incorporated into any network, \emph{making it highly general and scalable}.
Another benefit that our method should gain, in comparison with some prior methods~\cite{Gao2017, Kendall2018}, comes from the degradation itself, \emph{whose information may facilitate the image restoration process as well}~\cite{Zhang2018}.

We primarily focus on three image restoration tasks, \ie, inpainting, interpolation, and SISR (in which image blurring is also introduced), hence what we mean image restoration in the paper indicates the three tasks. Our main contributions are:
\begin{description}[font=$\bullet$]
\item [\quad] We propose a novel and general method to generalize off-the-shelf image restoration CNNs to succeed over a spectrum of image degradations levels.

\item [\quad] By encouraging high likelihood in the architecture, our method utilizes information from different degradation levels to facilitate the restoration process.

\item [\quad] Our method is computationally efficient (since the introduced overhead is small), easy to implement, and can be readily incorporated into many image restoration CNNs. 

\item [\quad] The effectiveness of our DL-Net is testified on a bunch of benchmark datasets: CelebA~\cite{Liu2015}, SUN397~\cite{Xiao2016}, Set-5~\cite{Bevilacqua2012}, Set-14~\cite{Zeyde2010_On}, and BSD-500~\cite{Martin2001}. The experimental results demonstrate that our DL-Net outperforms some prior state-of-the-arts in various test cases. 
\end{description}

\section{Related Work}\label{sec:rel}

\textbf{Image restoration.}
Typical restoration tasks include SISR, inpainting, denoising, deblurring, just to name a few. 
On the point that many denoising CNNs have assured satisfactory performance in multiple degradation settings~\cite{Zhang2017, Lefkimmiatis2018}, we opt to cover other critical tasks where the problem arises. 
We mainly focus on inpainting (also known as image completion or hole-filling), interpolation, and SISR.
Further, as will be introduced in Section~\ref{sec:dln}, we assume that some basic degradation settings are known to the image restoration model, and it is likely for many image restoration and SISR applications.

Image inpainting~\cite{Bertalmio2000}, a task of making visual predictions for missing regions, is required when human users aim to erase certain regions on an image.
Early inpainting solutions predict information of the missing regions by exploiting isophotes direction field~\cite{Bertalmio2000, Bertalmio2003} or texture synthesis technologies~\cite{Drori2003, Komodakis2006, Barnes2009}. 
Deep CNNs were later introduced to learn semantic content in an end-to-end manner.
Encoder-decoder-based architecture and adversarial learning are introduced to fill very large holes~\cite{Pathak2016, Iizuka2017, Yang2017, Yeh2017, Li2017, Yu2018}.
Despite impressive improvements, popular deep CNN-based methods inpaint with pre-defined region size, shape, and even location, and a deteriorating effect has been reported when multiple degradations exist~\cite{Gao2017}. 
This problem can be mitigated by bringing in some more holistic and local loss terms. 
Recently, Ren \etal\cite{Ren2015} and Liu \etal\cite{Liu2018} propose Shepard convolution and partial convolution for inpainting with irregular holes, respectively.
These studies of new loss terms and new design of convolutions are orthogonal to ours.

Image interpolation~\cite{Gao2017} is a similar task to inpainting, and it is sometimes also referred to as another type of inpainting (with Bernoulli masks).
Sparse coding networks are usually adopted to cope with it~\cite{Heide2015, Papyan2017}.
Some implicit prior captured by the CNN architecture itself is also explored~\cite{Ulyanov2018}.

Then it is SISR, the task of generating a high-resolution image given a low-resolution one.
Many classical SISR methods attempted to achieve the goal by preserving (example-based) neighborhood structures~\cite{Chang2004,  Fan2007, Bevilacqua2012, Timofte2013} or (dictionary-based) perceptual representations~\cite{Yang2008, Yang2010, Dong2011, Zhang2012, Timofte2013} in low-resolution and high-resolution images, such that some implicit mappings could be established between them.
Researchers have also tried pursuing approximated but explicit mappings from input to target subspaces directly~\cite{Dong2014, Gu2015}, and CNNs can be appropriate candidates for the mappings.
Being able to extract contextual information, networks with deep architectures showed promising SISR results~\cite{Kim2016, Mao2016, Tai2017, Sonderby2017, Hui2018}.
Nevertheless, it has also been reported that state-of-the-art SISR network models suffer from generalizing to scenarios with possible blurring~\cite{Zhang2018}.

\textbf{Likelihood and image priors.} A common ingredient of many non-deep image restoration solutions is to estimate the probability that one outcome is capable of generating the given input through some sort of degradations, \ie, the likelihood.
In principle, such a task is heavily ill-posed. 
Solving them would normally require prior knowledge on the input low-quality and output high-quality image subspaces, and learning-based methods are typically used to leverage them effectively. 
Generally, the likelihood and priors can be integrated together using the maximum a posteriori (MAP) or Bayesian frameworks~\cite{Freeman2000}. 
This line of research used to provide state-of-the-art results before the unprecedented success of deep learning. 

Early work advocates Gaussian process prior~\cite{Freeman2000}, Huber prior, sampled texture prior~\cite{Pickup2004}, edge prior~\cite{Tai2010}, etc.
Recently, generative-adversarial-network-derived~\cite{Goodfellow2014} image priors have gained great success~\cite{Pan2018, Sonderby2017}.
Our DL-Net exploits the likelihood and also image priors if necessary. 
Rather than adding them to the loss function, we reformulate their optimization procedure as a recursive module. In such a way, the functional of maximum likelihood can be readily incorporated into any given image restoration architecture.
Our method is closely related to deep image prior~\cite{Ulyanov2018}, in a sense that the methods both schematically suggest outputs that being able to reproduce the corresponding inputs and can be applied to various CNNs (see Section~\ref{sec:mdc} for discussions). Our method also lies in the category of MAP inference guided discriminative learning~\cite{Zhang2017_IR} and differential optimization architectures~\cite{agrawal2019differentiable}.   

\section{Deep Likelihood Network}\label{sec:dln}

Image restoration has been intensively studied for decades.
Before the advent of deep learning, conventional MAP-based methods hinged on the likelihood and image prior modeling had been widely applied and achieved state-of-the-art results on a variety of image restoration tasks.
In this work, we advocate the conventional MAP-based formulation and introduce its fidelity term to off-the-shelf image restoration CNNs, for handling multiple levels of degradations. 

In the following two subsections, we will first compare different problem formulations for image restoration and show that simply adding the fidelity term to a discriminative learning objective cannot effectively enhance an image restoration network under multiple degradation levels. 
Then, in Section~\ref{sec:our}, we will present our method that incorporates an MAP-inspired module with degradation information into the network architectures. 

\subsection{Problem Formulations: MAP \& Deep Learning}\label{sec:map}

In this work, we consider a group of image restoration tasks (include image inpainting, interpolation, and SISR) whose degradation process can be formulated as 
\begin{equation}\label{eq:3.0}
y = (Wx)\downarrow_t + \epsilon, 
\end{equation}
where $x \in \mathbb R^{ht\times wt \times c}$ and $y \in \mathbb R^{h\times w \times c}$ denote the clean and degraded images, respectively.
$\downarrow_t$ indicates the  downsampling operator (with an integer downsampling factor $t \geq 1$), $\epsilon$ is the additive white Gaussian noise with standard deviation $\sigma$, and $W$ is the task-specific degradation matrix.
For image inpainting and image interpolation, we have $Wx = M \odot x$, $t = 1$, and $\epsilon = 0$, where $\odot$ indicates the entry-wise multiplication of two tensors and $M$ denotes a binary mask.
For SISR, we have  $Wx = U \ast x$, where $\ast$ indicates the tensor convolution and $U$ denotes the degradation (\eg, blur) kernel.
In terms of \emph{multiple degradations} (or particularly multiple degradation levels in the paper), we aim to train a single model to handle a concerned image restoration task with a spectrum of degradation settings.

The MAP-based formulation of image restoration generally involves a fidelity term for modeling the likelihood of degradation and a regularization term for modeling image priors.
Given the degraded observation $y$ and the degradation setting (\ie, $W$, $t$, $\sigma$), the fidelity term can be written as,
\begin{equation}\label{eq:3.1}
\mathcal L(x,y) = \frac{1}{2\sigma^2}\|(Wx)\downarrow_t - y\|^2, 
\end{equation}
By further trading off against a regularization term $\mathcal R(x)$, the MAP problem can be formulated as,
\begin{equation}\label{eq:3.2}
\tilde{x} = \arg\min_x {\mathcal L}(x,y) + \alpha \mathcal R(x),
\end{equation}
where $\alpha \geq 0$ is the regularization parameter.

Since the degradation process is encoded in the fidelity term in Eqn.~(\ref{eq:3.1}), given the regularizer $\mathcal R(x)$ and an optimizer, the MAP framework is naturally flexible in handling multiple degradations.
However, most existing regularizers are hand-crafted, non-convex, and insufficient for characterizing natural image priors. 
On account of the non-convexity, the optimization algorithm often cannot find a satisfactory solution to Eqn. (\ref{eq:3.2})~\cite{Schmidt2014}.
All these problems make recent work resort to deep CNNs for improving restoration performance.

Given a specific degradation setting with parameters $W_0$, $t_0$, and $\sigma_0$, a deep learning model directly learns an input-output mapping from a training set of degraded and clean image pairs $\{ (x_i, y_i) \}_{i=1}^N$ to generate: $\hat{x}_i=f(y_i; \Theta_{W_0, t_0,\sigma_0})$.
In spite of their success~\cite{Dong2014, Pathak2016, Zhang2017}, the obtained models are often tailored to the specific degradation (with $W_0$, $t_0$, and $\sigma_0$) and generalize poorly to other degradation levels (with $W \neq W_0$, $t \neq t_0$, or $\sigma \neq \sigma_0$).
As such, we aim at developing a deep-learning-based image restoration method for handling multiple degradations. We hope it has the flexibility of MAP and the (accuracy and efficiency) merits of off-the-shelf deep networks.   
One specious method for this seems to add the fidelity term to the learning objective of CNNs.
Yet, as will be shown in Section~\ref{sec:ldn}, it does not succeed as expected.

\subsection{Likelihood in Learning Objective?}\label{sec:ldn}

It has been discussed in prior work~\cite{Gao2017,Zhang2018} that rigid-joint training supported by data augmentation provides limited assistance to generalizing the degradation settings.
The empirical results as will be shown in Section~\ref{sec:exp1} echo this claim.
In this subsection, we further show that the obtained performance of off-the-shelf networks cannot be effectively improved by adding the fidelity term to the learning objective either. 
Taking multiple degradations, an augmented training set is represented as $\{ (x_i, W_i, y_i) \}_{i=1}^{N'}$, where each sample is allowed to have its own degradations therefore we have $N'\gg N$ if desired in practice.
Using the augmented set, we can train an autoencoder to generate $\hat{x}_i = f(y_i; \Theta)$ for image restoration by minimizing the reconstruction loss $\mathcal{L}_{rec} \propto \sum_{i} \|x_i - \hat{x}_i\|^2$. The method is named Autoencoder (Joint) and some detailed explanations for our experimental settings will be given in Section~\ref{sec:exp1}. 

To incorporate the MAP formulation into deep learning solutions, we modify the fidelity term defined in Eqn.~(\ref{eq:3.1}) into $\mathcal L  = \frac{1}{2\sigma^2}\sum_i\|(W_i{\hat x_i})\downarrow_t - y_i\|^2$, which may entail  ``self-supervision''. 
The modified term can be combined with $\mathcal{L}_{rec}$, and the most direct combination is to use $\mathcal L_{rec}+\lambda \mathcal L$ as the objective function. 
One may expect such a scheme, dubbed Na\"ive Likelihood in this paper, to endow the obtained models some ability of handling multiple degradation levels.

\begin{table}[ht]
\begin{center}\resizebox{0.99\linewidth}{!}{
\begin{tabular}{|C{1.4in}|C{1.3in}|C{1.3in}|C{1.3in}|}
\hline
Method &  \tabincell{c}{$s=20$ \\ $l_1$ loss\,/\,$l_2$ loss\,/\,PSNR } &  \tabincell{c}{$s=20$,\, centered \\ $l_1$ loss\,/\,$l_2$ loss\,/\,PSNR } \\
\hline\hline
Autoencoder (Joint)~\cite{Gao2017}          & 0.0305\,/\,0.0040\,/\,30.87 &  0.0295\,/\,0.0035\,/\,31.50 \\
Na\"ive Likelihood ($\lambda=10^2$)        		    & 0.0304\,/\,0.0040\,/\,30.87 & 0.0295\,/\,0.0036\,/\,31.48  \\  
Na\"ive Likelihood ($\lambda=10^3$)        		    & 0.0296\,/\,0.0040\,/\,30.96 & 0.0287\,/\,0.0035\,/\,31.61  \\   
Na\"ive Likelihood ($\lambda=10^4$)        		    & 0.0258\,/\,0.0038\,/\,31.20 & 0.0248\,/\,0.0033\,/\,31.94  \\   
Na\"ive Likelihood ($\lambda=10^5$)        		    & 0.0248\,/\,0.0044\,/\,30.45 & 0.0235\,/\,0.0037\,/\,31.30  \\  \hline 
\end{tabular}}
\end{center}
\caption{Image inpainting results of autoencoder trained with $\mathcal L_{rec}$ and $\mathcal L_{rec} + \lambda\mathcal L$. Though slightly better results can be obtained (for $s=20$) using $\lambda=10^4$, it diminishes the PSNR under $s=30$ inevitably by 0.09dB.}
\label{tab:naive} \vskip -0.05in
\end{table}

\begin{figure}[t]
\begin{center}
\includegraphics[width=0.42\textwidth,trim={0.3in 0.8in 0.3in 1.1in},clip]{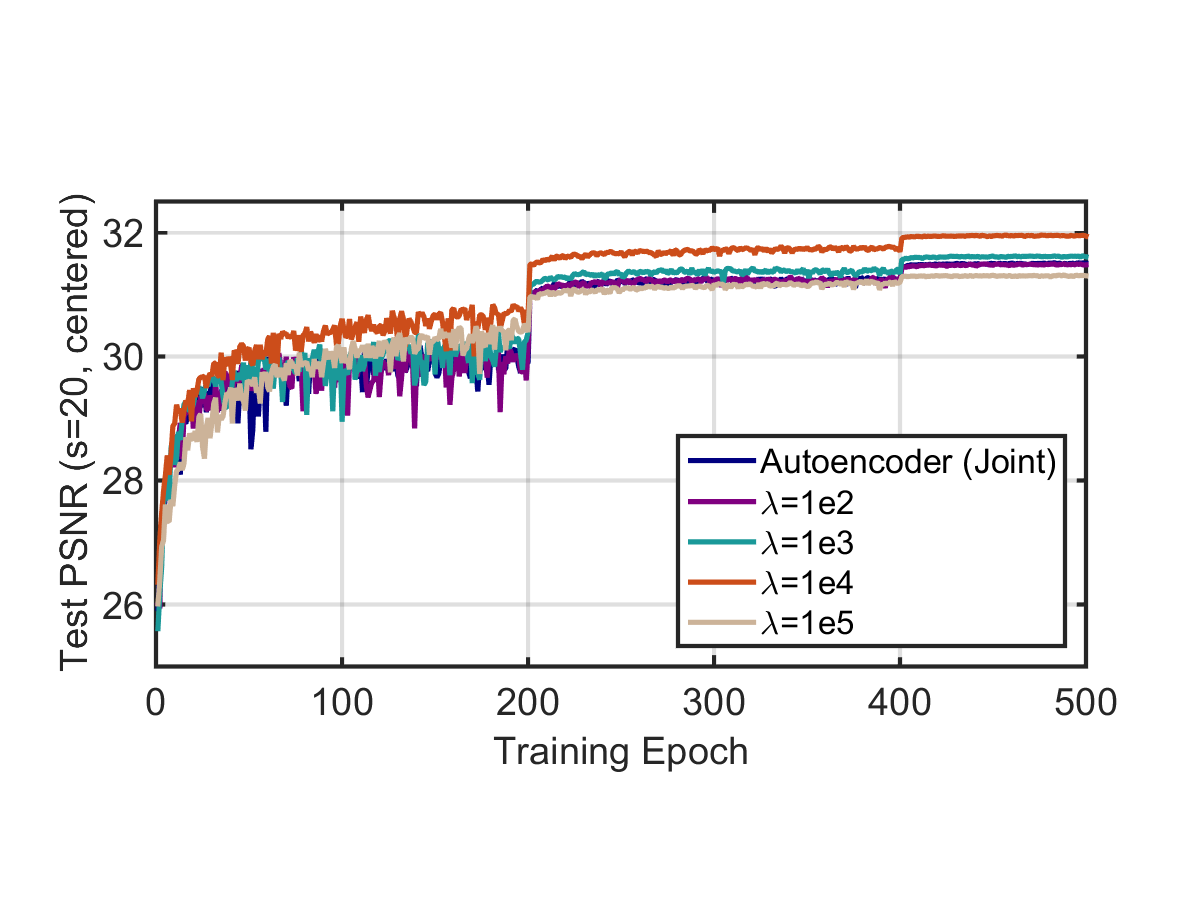}\vskip -0.1in
\caption{The training curves of ``Autoencoder (Joint)" and ``Na\"ive Likelihood" methods with various $\lambda$ values.} 
\label{fig:naive}
\end{center}
\vskip -0.15in
\end{figure}

To illustrate the effect of $\mathcal L$ in the objective, we present some quantitative results in Table~\ref{tab:naive}. 
Training curves ($s=20$, centered) are further provided in Figure~\ref{fig:naive}.
The experiment was performed on the image inpainting task (the symbol $s$ indicates the region size) and no adversarial loss is used here to guarantee reliable PSNRs for evaluating the restoration performance.
From Table~\ref{tab:naive} and Figure~\ref{fig:naive}, we can easily observe that, for most $\lambda$ values, the ``Na\"ive Likelihood" method always performs similarly when compared with ``Autoencoder (Joint)" which minimizes only $\mathcal L_{rec}$. 
Although $\lambda=10^4$ slightly improves the inpainting performance with $s=20$, we observed that, it was only because more attention was paid to the unmasked region, and \emph{the obtained model still failed in the inpainting setting with $s=30$}. 
More specifically, it achieved 26.04dB after retraining, which is even worse than Autoencoder (Joint): 26.13dB. 
As discussed in prior work~\cite{Zhang2018}, the major reason that such an incorporation of $\lambda\mathcal L$ in the objective function is less helpful in handling multiple degradation levels can be ascribed to the lack of degradation information when processing images in deep networks. 

\subsection{Likelihood Assured by A Recursive Module}\label{sec:our}

\begin{figure}[t]
\begin{center}
\subfloat[]{\label{fig:2a}
\includegraphics[height=0.25\textwidth]{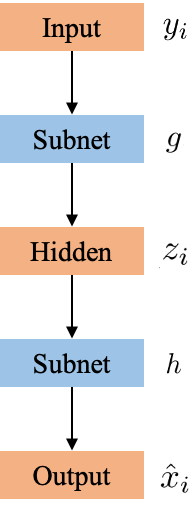}} \hskip 0.08in
\subfloat[]{\label{fig:2b}
\includegraphics[height=0.25\textwidth]{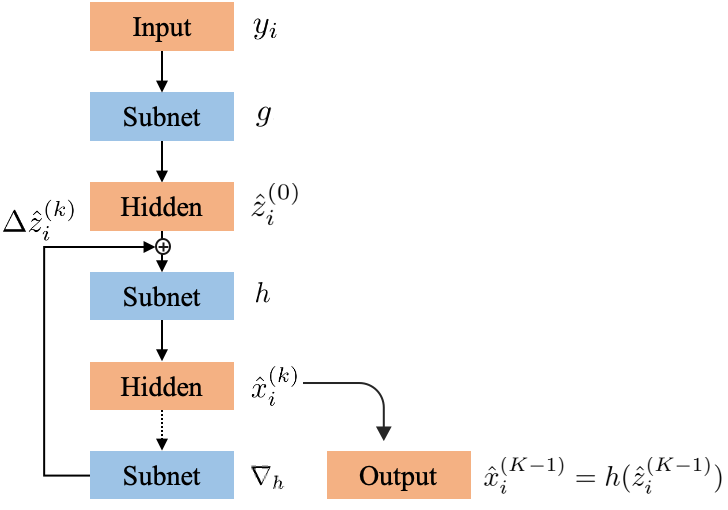}} 
\vskip -0.05in
\caption{Illustration of (a): the reference network and (b) our DL-Net. In each recursive step of DL-Net, a tensor $\Delta \hat z_i^{(k)}$ is calculated and added for approaching the final $\hat z_i$. The restoration result $\hat{x}_i$ can be iteratively estimated but only the one at iteration $K$ will be taken as the final output.}
\label{fig:2}
\end{center}
\vskip -0.22in
\end{figure}

In this subsection, we present our method for generalizing image restoration networks to handling a spectrum of degradation settings. 
As depicted in Figure~\ref{fig:2a}, given an off-the-shelf image restoration network (\ie, the reference architecture), the output of an intermediate layer is denoted by $z_i=g({y_i})$ and we also denote $\hat{x}_i = h(z_i)$, which means the reference network $f = h\circ g$ is regarded as the composition of two functions, or in other words two sub-networks.   

Partially inspired by deep image prior~\cite{Ulyanov2018}, we substitute $\hat{x}$ with $h(z)$ in the fidelity term and eliminate the explicit regularization term, since some implicit priors can be characterized by the sub-network $h$. 
In contrast to~\cite{Ulyanov2018}, we assume there exists a unified $h$ suitable to all possible inputs, but, for each of them, a specific $z$ should be learned with an optimization algorithm.  
In this regard, the formulation in Eqn.~(\ref{eq:3.2}) can further be rewritten as
\begin{equation}\label{eq:3.4}
\hat{x}_i = h(\hat z_i), \quad \hat z_i = \arg\min_z \|(W_ih(z))\downarrow_t - {y_i}\|^2.
\end{equation}

Since the reference $f$ is differentiable, $h$ and the whole objective function in~(\ref{eq:3.4}) are both differentiable w.r.t. $z$. 
Therefore, we can simply use a stochastic gradient descent algorithm to pursue $\hat{z}_i$.
Here we choose ADAM~\cite{Kingma2015} for this task, and we keep all its hyper-parameters except for the learning rate as default, see Eqn.~(\ref{eq:3.4.1}) for more details.
Such a computation procedure of using ADAM to update and pursue $\hat{z}_i$ can be cast into a recursive module and incorporated directly into any image restoration architecture.
That is, at the $k$-th iteration, $\Delta \hat z_i^{(k)}$ is calculated in the network and added to the previous estimation to obtain $\hat z_i^{(k)}=\hat z_i^{(k-1)} +  \Delta \hat z_i^{(k)}$. 
In principle, as long as $\hat z_i^{(0)} = g(y_i)$ is set, an update of the estimation leads to a superior model to the reference~\cite{Combes2018}.
After $K$ rounds of recursive computations, $\mathcal L_{rec}$ can be evaluated by calculating the norm of the difference between the final restoration result $\hat x_i^{(K-1)}=h(\hat z_i^{(K-1)})$ and the corresponding groundtruth $x_i$. 
Note that, although a set of $\hat{x}_i^{(k)}$, for $k=0,\ldots,K-1$, are sequentially calculated for $y_i$, only the last one (\ie, $\hat{x}_i^{(K-1)})$ will be taken as the final output. 

The proposed method is dubbed deep likelihood network (DL-Net) in the paper. Figure~\ref{fig:2b} illustrates the modifications invoked by our DL-Net, on the basis of the reference architecture.
Considering that $\nabla \hat z^{(k)}$ (\ie, the gradient of the objective function given in Eqn.~(\ref{eq:3.4}) w.r.t. the estimation $\hat z_i^{(k-1)}$ at the $k$-th step) is involved in the representation of $\Delta \hat z_i^{(k)}$ and $\hat{x}_i^{(K-1)}$, the computation of higher-order gradients is required in the training process, and automatic differentiation in current deep learning frameworks (\eg, TensorFlow and PyTorch) can handle it efficiently.

In comparison with the reference network, all our modifications (cf Figure~\ref{fig:2a} and~\ref{fig:2b}) occur on the basis of $h$. 
A feedback connection is established for estimating $\hat{z}_i$ with $\Delta \hat z_i^{(k)}$, and the computations involved for deriving $\Delta \hat z_i^{(k)}$ from the gradient $\nabla \hat z_i^{(k)}$ are:
\begin{equation}\label{eq:3.4.1}
\begin{split}
m^{(k)} &= \beta_1 m^{(k-1)}+(1-\beta_1)\nabla z_i^{(k)}, \\ 
v^{(k)} &= \beta_2 v^{(k-1)}+(1-\beta_2)\nabla z_i^{(k)}\cdot\nabla z_i^{(k)}, \\
\hat{m}^{(k)} &= m^{(k)}/(1-\beta_1), \\
\hat{v}^{(k)} &= v^{(k)}/(1-\beta_2), \\
\Delta \hat z_i^{(k)} &= -\gamma\hat{m}^{(k)}/(\sqrt{\max(\hat{v}^{(k)}, 10^{-16})}).
\end{split}
\end{equation}

The crux of our DL-Net is to incorporate an MAP-inspired module that is able to assist $h$ for disentangling the effects of different degradations.
It enforces the outputs to achieve lower  $\mathcal L$ iteratively, in a similar vein to AffGAN~\cite{Sonderby2017}.
Our method may as well be employed in a bunch of classical learning-based methods where deep networks are left-off. 
Moreover, it is natural to further incorporate other desirable priors into~(\ref{eq:3.4}).
One extra benefit that such an MAP-inspired method should have, as emphasized in~\cite{Zhang2018}, comes along with some insightful knowledge extracted from the degradation process, which might be critical for facilitating the learning. 
For instance, when the location, size, and shape of the inpainting regions are known, inpainting networks are likely to learn automatically paying more attention to these regions~\cite{Yu2018}.

Since a gradient-descent-derived module is utilized, two more hyper-parameters will be introduced: the total number of gradient descent steps $K$ and the learning rate $\gamma$ for~(\ref{eq:3.4}). 
We fixed $K = 5$ and $\gamma = 10^{-3}$ unless otherwise clarified in the paper. 
Obviously, the depth of $h$ hence has a major impact on the computational cost of our DL-Net.
We found that a light $h$ with one single convolution layer only (along with a nonlinear activation layer picked up from its previous layer) worked well enough in practice, so the extra running time could be negligible in comparison to $g$ which is much deeper. 
That is, $z_i$ was chosen as the last hidden representation (before ReLU) from the reference network and thus the increase of computational cost is small.
The settings were adopted across all the experiments in this work.

\section{Experimental Results}\label{sec:exp}

\subsection{Image Inpainting and Interpolation}~\label{sec:exp1}

We will analyze the performance of different models on image inpainting and interpolation together in this subsection.

\textbf{Network architectures.} Following prior work~\cite{Gao2017}, we first introduce inpainting networks for hallucinating visual contents within square blocks whose size and location vary randomly in given images, and a similar pipeline will be applied also to image interpolation.
As a baseline model, the popular encoder-decoder-based architecture~\cite{Pathak2016, Gao2017, Li2017} (a.k.a., autoencoder) was chosen. We directly adapted the open-source Torch implementation from Gao and Grauman~\cite{Gao2017}. 
See Section~\ref{sec:netarc} for more details.  

\begin{table*}[t]
\begin{center}\resizebox{0.96\linewidth}{!}{
\begin{tabular}{|C{1.65in}|C{1.35in}|C{1.35in}|C{1.35in}|C{1.35in}|}
\hline
Method & \tabincell{c}{ $s=10$ \\ $l_1$ loss\,/\,$l_2$ loss\,/\,PSNR } &  \tabincell{c}{$s=20$ \\ $l_1$ loss\,/\,$l_2$ loss\,/\,PSNR } &  \tabincell{c}{$s=30$ \\ $l_1$ loss\,/\,$l_2$ loss\,/\,PSNR }  &  \tabincell{c}{$s=20$,\, centered \\ $l_1$ loss\,/\,$l_2$ loss\,/\,PSNR } \\
\hline\hline
Autoencoder (Default)      & 0.2556\,/\,0.1269\,/\,15.49 & 0.2482\,/\,0.1299\,/\,15.38 & 0.2600\,/\,0.1733\,/\,14.17 & 0.2470\,/\,0.1226\,/\,15.62 \\
Autoencoder ($s=20$,\, centered)     & 0.2116\,/\,0.0951\,/\,17.16 & 0.2258\,/\,0.1402\,/\,15.67 & 0.2616\,/\,0.2321\,/\,13.19 & 0.0246\,/\,0.0031\,/\,32.27 \\
Autoencoder (Joint)~\cite{Gao2017}               & 0.0227\,/\,0.0015\,/\,35.00 & 0.0305\,/\,0.0040\,/\,30.87 & 0.0494\,/\,0.0118\,/\,26.13 & 0.0295\,/\,0.0035\,/\,31.50 \\
On-Demand Learning~\cite{Gao2017}                  & 0.0230\,/\,0.0015\,/\,34.93 & 0.0307\,/\,0.0040\,/\,30.85 & 0.0496\,/\,0.0118\,/\,26.12 & 0.0299\,/\,0.0036\,/\,31.43 \\
Multi-Task Learning~\cite{Kendall2018} 		    & 0.0178\,/\,0.0010\,/\,36.65 & 0.0263\,/\,0.0036\,/\,31.37 & 0.0470\,/\,0.0118\,/\,26.11 & 0.0253\,/\,0.0032\,/\,32.06 \\
DL-Net (ours)               						    & \textbf{0.0129}\,/\,\textbf{0.0008}\,/\,\textbf{38.47} & \textbf{0.0214}\,/\,\textbf{0.0033}\,/\,\textbf{32.03} & \textbf{0.0413}\,/\,\textbf{0.0110}\,/\,\textbf{26.54} & \textbf{0.0205}\,/\,\textbf{0.0028}\,/\,\textbf{32.82}   \\ \hline
\end{tabular}}
\end{center}
\caption{Image inpainting with multiple degradations on CelebA: our method compared with the baseline and competitive methods. Apparently, our DL-Net consistently outperforms the others in all test cases.} \vskip -0.2in
\label{tab:celeba_ii}
\end{table*}

\begin{table*}[t]
\begin{center}\resizebox{0.96\linewidth}{!}{
\begin{tabular}{|C{1.65in}|C{1.35in}|C{1.35in}|C{1.35in}|C{1.35in}|}
\hline
Method & \tabincell{c}{ $r=15\%$ \\ $l_1$ loss\,/\,$l_2$ loss\,/\,PSNR } &  \tabincell{c}{$r=45\%$ \\ $l_1$ loss\,/\,$l_2$ loss\,/\,PSNR } &  \tabincell{c}{$r=75\%$ \\ $l_1$ loss\,/\,$l_2$ loss\,/\,PSNR }  &  \tabincell{c}{$r\in[0\%,75\%]$,\, average \\ $l_1$ loss\,/\,$l_2$ loss\,/\,PSNR } \\
\hline\hline
Autoencoder (Default)      & 0.0924\,/\,0.0159\,/\,24.64 & 0.0527\,/\,0.0062\,/\,28.59 & 0.0565\,/\,0.0095\,/\,26.78 & 0.0705\,/\,0.0112\,/\,26.67 \\
Autoencoder (Joint)~\cite{Gao2017}               & 0.0256\,/\,0.0017\,/\,34.36 & 0.0347\,/\,0.0036\,/\,31.07 & 0.0597\,/\,0.0104\,/\,26.32 & 0.0345\,/\,0.0037\,/\,31.68 \\
On-Demand Learning~\cite{Gao2017}                  & 0.0251\,/\,0.0016\,/\,34.50 & 0.0343\,/\,0.0035\,/\,31.15 & 0.0589\,/\,0.0102\,/\,26.41 & 0.0340\,/\,0.0036\,/\,31.78 \\
Multi-Task Learning~\cite{Kendall2018} 		    & 0.0216\,/\,0.0013\,/\,35.58 & 0.0335\,/\,0.0034\,/\,31.24 & 0.0619\,/\,0.0110\,/\,26.05 & 0.0327\,/\,0.0035\,/\,32.22 \\
DL-Net (ours)               						    & \textbf{0.0138}\,/\,\textbf{0.0007}\,/\,\textbf{38.35} & \textbf{0.0240}\,/\,\textbf{0.0023}\,/\,\textbf{33.09} & \textbf{0.0488}\,/\,\textbf{0.0083}\,/\,\textbf{27.42} & \textbf{0.0233}\,/\,\textbf{0.0024}\,/\,\textbf{34.51}   \\ \hline
\end{tabular}}
\end{center}
\caption{Image interpolation with multiple degradations on CelebA: our method ($\gamma=10^{-2}$) compared with the baseline and competitive methods. Apparently, our DL-Net consistently outperforms the others in all test cases.} \vskip -0.1in
\label{tab:celeba_it}
\end{table*}

\begin{table*}[h]
\begin{center}\resizebox{0.95\linewidth}{!}{
\begin{tabular}{|C{1.6in}|C{1.48in}|C{1.48in}|C{1.48in}|C{1.48in}|}
\hline
Method & \tabincell{c}{ $s=10$ \\ $l_1$ loss\,/\,$l_2$ loss\,/\,PSNR } &  \tabincell{c}{$s=20$ \\ $l_1$ loss\,/\,$l_2$ loss\,/\,PSNR } &  \tabincell{c}{$s=30$ \\ $l_1$ loss\,/\,$l_2$ loss\,/\,PSNR }  &  \tabincell{c}{$s=20$,\, centered \\ $l_1$ loss\,/\,$l_2$ loss\,/\,PSNR } \\
\hline\hline
Autoencoder (Default)      & 0.2407\,/\,0.1098\,/\,16.19 & 0.2297\,/\,0.1088\,/\,16.19 & 0.2626\,/\,0.1709\,/\,14.27 & 0.2303\,/\,0.1033\,/\,16.38 \\
Autoencoder ($s=20$,\, centered)     & 0.1926\,/\,0.0835\,/\,17.73 & 0.1845\,/\,0.1104\,/\,16.79 & 0.2400\,/\,0.2141\,/\,13.47 & 0.0485\,/\,0.0132\,/\,25.84 \\
Autoencoder (Joint)~\cite{Gao2017}               & 0.0299\,/\,0.0031\,/\,32.18 & 0.0438\,/\,0.0098\,/\,27.16 & 0.0702\,/\,0.0235\,/\,23.20 & 0.0444\,/\,0.0102\,/\,26.95 \\
On-Demand Learning~\cite{Gao2017}                  & 0.0311\,/\,0.0032\,/\,31.94 & 0.0450\,/\,0.0101\,/\,27.04 & 0.0711\,/\,0.0238\,/\,23.15 & 0.0456\,/\,0.0105\,/\,26.83 \\
Multi-Task Learning~\cite{Kendall2018} 		    & 0.0245\,/\,0.0025\,/\,33.20 & 0.0389\,/\,0.0093\,/\,27.48 & 0.0662\,/\,0.0231\,/\,23.31 & 0.0395\,/\,0.0097\,/\,27.24 \\
DL-Net (ours)               						    & \textbf{0.0175}\,/\,\textbf{0.0019}\,/\,\textbf{34.71} & \textbf{0.0329}\,/\,\textbf{0.0090}\,/\,\textbf{27.73} & \textbf{0.0608}\,/\,\textbf{0.0229}\,/\,\textbf{23.39} & \textbf{0.0337}\,/\,\textbf{0.0094}\,/\,\textbf{27.44}   \\ \hline
\end{tabular}}
\end{center}
\caption{Image inpainting with multiple degradations on SUN397: our method compared with the baseline and competitive methods.}\vskip -0.1in
\label{tab:sun397_ii}
\end{table*}

\begin{table*}[h]
\begin{center}\resizebox{0.95\linewidth}{!}{
\begin{tabular}{|C{1.6in}|C{1.48in}|C{1.48in}|C{1.48in}|C{1.48in}|}
\hline
Method & \tabincell{c}{ $r=15\%$ \\ $l_1$ loss\,/\,$l_2$ loss\,/\,PSNR } &  \tabincell{c}{$r=45\%$ \\ $l_1$ loss\,/\,$l_2$ loss\,/\,PSNR } &  \tabincell{c}{$r=75\%$ \\ $l_1$ loss\,/\,$l_2$ loss\,/\,PSNR }  &  \tabincell{c}{$r\in[0\%,75\%]$,\, average \\ $l_1$ loss\,/\,$l_2$ loss\,/\,PSNR } \\
\hline\hline
Autoencoder (Default)      & 0.1575\,/\,0.0390\,/\,20.34 & 0.0717\,/\,0.0119\,/\,25.87 & 0.0856\,/\,0.0199\,/\,23.71 & 0.1213\,/\,0.0351\,/\,23.10 \\
Autoencoder (Joint)~\cite{Gao2017}               & 0.0348\,/\,0.0032\,/\,31.83 & 0.0488\,/\,0.0071\,/\,28.28 & 0.0820\,/\,0.0187\,/\,23.96 & 0.0480\,/\,0.0072\,/\,29.06 \\
On-Demand Learning~\cite{Gao2017}                  & 0.0342\,/\,0.0031\,/\,31.98 & 0.0480\,/\,0.0069\,/\,28.39 & 0.0809\,/\,0.0184\,/\,24.05 & 0.0472\,/\,0.0070\,/\,29.17 \\
Multi-Task Learning~\cite{Kendall2018} 		    & 0.0292\,/\,0.0025\,/\,32.98 & 0.0465\,/\,0.0068\,/\,28.46 & 0.0845\,/\,0.0195\,/\,23.74 & 0.0451\,/\,0.0069\,/\,29.58 \\
DL-Net (ours)               						    & \textbf{0.0179}\,/\,\textbf{0.0014}\,/\,\textbf{35.28} & \textbf{0.0368}\,/\,\textbf{0.0054}\,/\,\textbf{29.53} & \textbf{0.0722}\,/\,\textbf{0.0166}\,/\,\textbf{24.56} & \textbf{0.0343}\,/\,\textbf{0.0054}\,/\,\textbf{31.38}   \\ \hline
\end{tabular}}
\end{center}
\caption{Image interpolation with multiple degradations on SUN397: our method ($\gamma=10^{-2}$) compared with the baseline and competitive methods.}\vskip -0.3in
\label{tab:sun397_it}
\end{table*}

\textbf{Training and test samples.} Same with Gao and Grauman~\cite{Gao2017}, we evaluated inpainting and interpolation models on two datasets: CelebA~\cite{Liu2015} and SUN397~\cite{Xiao2016}.
For CelebA, the first 100,000 images were used to form the training set. 
Some other 2000 images were randomly chosen and split into a validation set and a test set, each contains 1000 uncorrupted images.
For SUN397, we similarly had 100,000 images for training, and we also had 1000 uncorrupted images each for validation and testing. 
Input images were uniformly rescaled to $64\times 64$, thus the encoder would output feature maps with a spatial size of $4\times4$.
Randomness was introduced into the test samples owing to pixel removal at possibly any location, thus in order to ensure that different models were tested on the same set, we utilized $10\times$ more samples by degenerating each original test image with $10$ feasible degradations and save all these combinations (of corrupted images and the ground-truth) locally for testing. That is, all the models were tested on the same sample pairs.

\textbf{Training process.} We initialized the channel-wise fully-connected layer with a random Gaussian distribution and all convolutional layers with the ``MSRA'' method~\cite{He2015}.
In our experiments, $10^{-4}$ weight decay and ADAM~\cite{Kingma2015} with an initial learning rate of $10^{-3}$  were adopted to optimize:
\begin{equation}\label{eq:3.5}
\mathcal L_{rec}=\frac{1}{whN}\sum^{N-1}_{i=0} \|\hat{x}_{i}-x_i\|^2
\end{equation}
for obtaining reference models.
Training batch size was set to be 100.
To keep in line with prior work, we report $l_1$, $l_2$ loss and the PSNR index on our test set for evaluating performance.
Images were numerically converted into the floating-point format and scaled to $[-1, 1]$ for calculating these values for evaluation. 
We cut the learning rate by $10\times$ every 100 epochs due to no improvement observed on the held-out validation set.
The reference inpainting model was trained to fill square holes with a spatial size of $32\times32$ (i.e., $s=32$), and the reference interpolation model was trained to restore images on which 75\% of the pixels were removed (i.e., $r=75\%$).
After training for 250 epochs, these reference models reached plateau and we achieves a PSNR of 24.45dB and 26.78dB on CelebA, for inpainting with size $s=32$ and interpolation with $r=75\%$, respectively.

\textbf{Main results.} Despite the decent results on a presumed and fixed degradation setting, the obtained reference models failed when evaluated on other inpainting and interpolation settings that were not specifically trained in, and might be technically easier.
For example, a PSNR of only 15.62dB was achieved when we moderately adjusted the test inpainting size $s$ to $20$.
When the location constraint was further relaxed, it degraded to 15.38dB, which is by no means satisfactory in practice. 
Detailed quantitative results can be found in Table~\ref{tab:celeba_ii} and~\ref{tab:celeba_it}~\footnote{For results on SUN397, please see Table~\ref{tab:sun397_ii} and~\ref{tab:sun397_it}.}.
This problem has been comprehensively studies in~\cite{Gao2017}, so we followed their experimental settings and let $s$ and $r$ vary in $\{1,\ldots,30\}$ and $[0\%,75\%]$, respectively.
We let the inpainting regions shift randomly in a range of $[-10,10]$ pixels  around the centroid of images where most meaningful content exists.~\footnote{Note that as one of the few effective methods, our DL-Net generalizes also to irregular holes, we choose the above settings mostly to make fair comparisons with previous work~\cite{Gao2017, Kendall2018}.}

Typically, a rigid-joint training strategy~\cite{Pathak2016, Gao2017} might mitigate the problem and save the image restoration performance to some extent. 
It attempts to minimize the reconstruction loss over a variety of degradation settings simultaneously in training. 
Experimental results demonstrated that such a method led to an average PSNR of 32.40dB for inpainting and 31.68dB for interpolation, over all the degradation settings on our test sets. 
Though the learning problem seems more complex, the model converged reasonably fast by fine-tuning from reference models instead of training from scratch.
The training process took 500 epochs for both tasks, and the learning rate was cut by $10\times$ every 200 epochs.
We evaluated the $l_1$, $l_2$ loss and PSNR in specific degradation settings and summarized all the results on CelebA in Table~\ref{tab:celeba_ii} and~\ref{tab:celeba_it}.
(See Table~\ref{tab:sun397_ii} and~\ref{tab:sun397_it} for results on SUN397.) Models were trained with 100,000 images randomly sampled from the dataset and tested on some other 1000 images, just as in~\cite{Gao2017}.

\begin{figure*}[t]
\begin{center}
\includegraphics[width=0.84\textwidth]{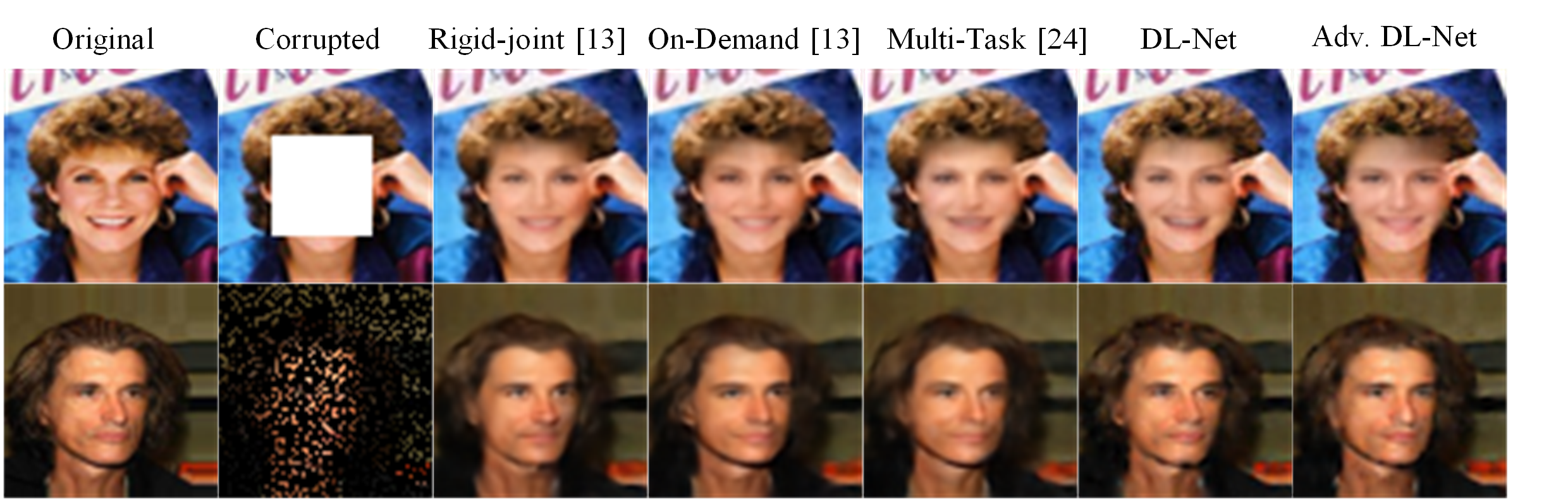}
\vskip -0.1in
\caption{Qualitative comparisons between our method and competitors. See for example the eyebrow and the mouth of the woman for their difference. The inpainting region is slightly biased to the left bottom of the images. These images are not cherry-picked, and better zoomed in for more details.}
\vskip -0.15in
\label{fig:celeba_ii_it_quality}
\end{center}
\end{figure*}

We then fine-tuned with our DL-Net method similarly and tested the obtained models.
As shown in Table~\ref{tab:celeba_ii} and~\ref{tab:celeba_it}, our DL-Net models gained significantly better results in comparison to rigid-joint training on all test cases.
We also compared our method with the state-of-the-art solutions dealing with the same problem in the literature, including those using multi-task learning and on-demand learning~\cite{Gao2017}.
Being aware of a very recent progress on multi-task learning using uncertainty to weigh losses~\cite{Kendall2018}, we tried adapting it to our image restoration tasks, for which each degradation level~\footnote{For inpainting, each $s$ was regarded as a particular level while for image interpolation, $r\in$[2.5($a$-1)\%, 2.5$a$\%) with a positive integer $a$ was regarded as one level. We also tried other possibilities but never got better results.} was treated as a subtask and a weight in the learning objective was introduced accordingly.
To be more specific, we had 30 extra weights to learn for both multi-task image inpainting and interpolation.
The on-demand learning method was configured exactly the same as suggested in the paper.
As expected, the two methods worked well on multiple degradations, outperforming the aforementioned joint training strategy in most cases.
However, they never surpass our DL-Net method in the context of restoration performance and PSNRs. 

For qualitative comparisons, see Figure~\ref{fig:celeba_ii_it_quality}.
It is also straightforward to absorb an adversarial loss in our DL-Net, either by introducing an adversarial prior $\mathcal R_{x}$ in Eqn.~(\ref{eq:3.4}) or adding it directly to $\mathcal L_{rec}$, we adopted the latter and illustrate its results in Figure~\ref{fig:celeba_ii_it_quality} as well.

\subsection{More Analyses}\label{sec:more_exp}
Our experimental settings in the previous subsection, including configurations on the training and test sets, CNN architectures, and loss terms for image inpainting and interpolation mostly comply with those in prior related work~\cite{Gao2017}.
Note that some different settings might also be popular in the task of image inpainting.
For example, since only part of the original images is missing, one may consider using $ I_x + (1-M) \odot \hat{I}_y$ rather than using $\hat{I}_y$ directly as the reconstruction result.
We compared our DL-Nets with competitive models in such a setting, and the restoration results were evaluated and reported in Table~\ref{tab:celeba_ii_combined} and~\ref{tab:celeba_it_combined}.
Apparently, our method still outperforms its competitors on both tasks.

\begin{table}[ht]
\begin{center}\resizebox{0.97\linewidth}{!}{
\begin{tabular}{|C{1.35in}|C{1.25in}|C{1.25in}|}
\hline
Method & \tabincell{c}{$s=20$ \\ $l_1$ loss\,/\,$l_2$ loss\,/\,PSNR } &  \tabincell{c}{$s=30$ \\ $l_1$ loss\,/\,$l_2$ loss\,/\,PSNR }   \\
\hline\hline
Autoencoder (Joint)~\cite{Gao2017}          & 0.0117\,/\,0.0031\,/\,32.37 & 0.0332\,/\,0.0109\,/\,26.52  \\   
On-Demand Learning~\cite{Gao2017}                  & 0.0117\,/\,0.0031\,/\,32.37 & 0.0331\,/\,0.0110\,/\,26.52  \\  
Multi-Task Learning~\cite{Kendall2018}  		    & 0.0117\,/\,0.0031\,/\,32.33 & 0.0341\,/\,0.0113\,/\,26.35  \\  
DL-Net (ours)                 						 & \textbf{0.0117}\,/\,\textbf{0.0031}\,/\,\textbf{32.42} & \textbf{0.0328}\,/\,\textbf{0.0108}\,/\,\textbf{26.62}    \\ \hline  
\end{tabular}}
\end{center}
\caption{Image inpainting with multiple degradations on CelebA: our DL-Net compared with competitors using $ I_x + (1-M) \odot \hat{I}_y$.}
\label{tab:celeba_ii_combined}
\vskip -0.18in
\end{table}

\begin{table}[ht]
\begin{center}\resizebox{0.97\linewidth}{!}{
\begin{tabular}{|C{1.35in}|C{1.25in}|C{1.25in}|}
\hline
Method & \tabincell{c}{$r=45\%$ \\ $l_1$ loss\,/\,$l_2$ loss\,/\,PSNR } &  \tabincell{c}{$r=75\%$ \\ $l_1$ loss\,/\,$l_2$ loss\,/\,PSNR }   \\
\hline\hline
Autoencoder (Joint)~\cite{Gao2017}          & 0.0204\,/\,0.0027\,/\,32.31 & 0.0512\,/\,0.0098\,/\,26.61  \\   
On-Demand Learning~\cite{Gao2017}                  & 0.0203\,/\,0.0027\,/\,32.35 & 0.0505\,/\,0.0096\,/\,26.70  \\  
Multi-Task Learning~\cite{Kendall2018}  		    & 0.0204\,/\,0.0027\,/\,32.27 & 0.0532\,/\,0.0103\,/\,26.34  \\   
DL-Net (ours)                 						 & \textbf{0.0171}\,/\,\textbf{0.0021}\,/\,\textbf{33.52} & \textbf{0.0439}\,/\,\textbf{0.0080}\,/\,\textbf{27.54}    \\ \hline  
\end{tabular}}
\end{center}
\caption{Image interpolation with multiple degradations on CelebA: DL-Net compared with competitors using $ I_x + (1-M) \odot \hat{I}_y$.}\vskip -0.15in
\label{tab:celeba_it_combined}
\end{table}

We know that there might exist some other choices for the formulation of $\mathcal L_{rec}$ for training inpainting networks, and our method directly generalizes to those cases.
In addition to the one derived in Eqn.~(\ref{eq:3.5}), some researchers~\cite{Pathak2016, Yu2018} also proposed to enlarge the inpainting blocks by 7 pixels during training and penalize $10\times$ more on the boundary regions for encouraging perceptual consistency.
As shown in Table~\ref{tab:celeba_ii_apx}, the superiority of our DL-Net holds. 

One may observe and consider it a bit surprising that our method even achieves better results than specifically trained references on certain degradation levels.
For instance, ``Autoencoder ($s=20$, centered)'' is slightly worse than DL-Net even when tested with centered $20\times 20$ regions.
Also, the ``Autoencoder (Default)'' interpolation model does not depict superior performance to our DL-Net when 75\% of the pixels are indeed removed. 
We believe this is partially because some insightful knowledge extracted from the size and location information of the blank regions facilitates the image restoration process~\cite{Zhang2018}. 
Two other plausible explanations are training samples of benefit from more flexible degradations and the probably enlarged network capacity owing to a recursive $h$.

\begin{figure}[t]
\begin{center}
\includegraphics[width=0.43\textwidth,trim={0.3in 0.7in 0.3in 1.05in},clip]{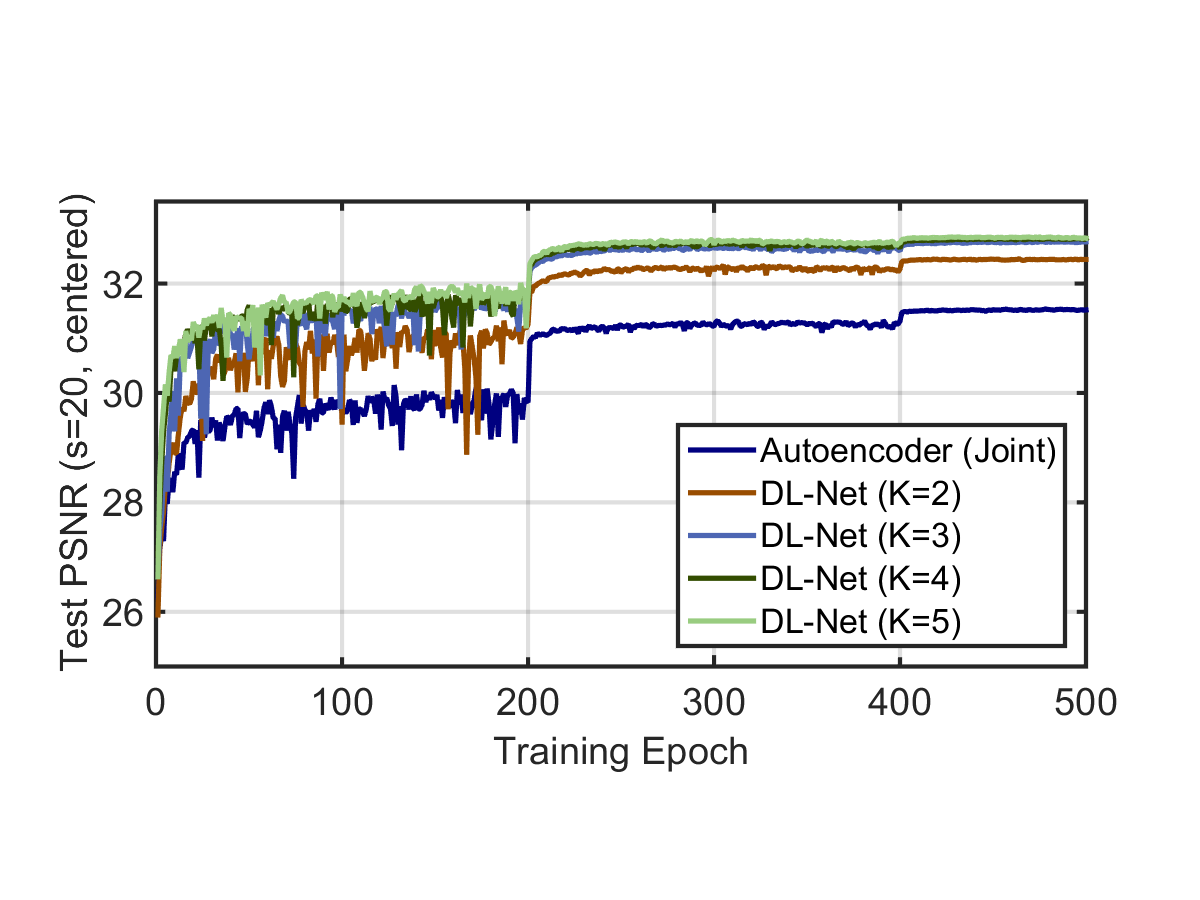}\vskip -0.15in
\caption{The training curves of a rigid-joint learning method and our DL-Net with various $K$. Note that the rigid-joint training can be considered as a special case of our DL-Net with $K=1$.}\vskip -0.2in
\label{fig:curve_dlk}
\end{center}
\end{figure}

We designed deeper reference networks and test if it was the capacity increase that helped.
Stacking four extra convolutional layers interlaced with ReLUs to the end (before the output layer), or putting them in the beginning, or allocating two on each side, we had three networks with higher capacities.
We trained them as training our ``Autoencoder ($s=20$,\, centered)'' model and tested them with centered $20\times20$ regions. 
Their average PSNRs were 32.21dB, 32.44dB, and 32.35dB, respectively, showing the increased capacity helped only to a limited extent.
Experiments were also conducted to demonstrate how the performance of our method varies with $K$.
Note that the rigid-joint training can be considered as a special case of DL-Net with $K=1$, we illustrate its training curve together with those of DL-Nets ($K\in\{2,3,4,5\}$) in Figure~\ref{fig:curve_dlk}. 
Apparently, larger $K$ indicates faster convergence and suggests better restorations.

\textbf{Overhead.} We evaluated CPU/GPU runtime of these DL-Nets with various $K$ values and summarize the results in Table~\ref{tab:runtime}.
It can be noticed that the introduced computational overhead is very small (at most 1.4$\times$ for $K=5$ on GPU)
The number of learnable parameters is also reported. 
Since our method utilizes a recursive module, it never brings in more learnable parameters.
All models discussed above are tested with the open-source TensorFlow framework~\cite{Abadi2016} on an Intel Xeon E5 CPU and an NVIDIA Titan X GPU. 

\begin{table}[ht]
\begin{center}\resizebox{0.97\linewidth}{!}{
\begin{tabular}{|C{1.25in}|C{1.0in}|C{1.0in}|C{0.70in}|}
\hline
Method & CPU Run-time (s)  &  GPU Run-time (s)  & \#Parameters   \\
\hline\hline
Autoencoder (Joint)~\cite{Gao2017}          & 3.824 & 1.330 &  \\   
DL-Net (K=2)                 				 & 4.346 & 1.595 &  \\
DL-Net (K=3)                 				 & 4.747 & 1.630 &  $\sim 2.9$M\\
DL-Net (K=4)                 				 & 5.255 & 1.747 &  \\
DL-Net (K=5)                 				 & 5.884 & 1.876 &   \\ \hline  
\end{tabular}}
\end{center}
\caption{Compare CPU/GPU runtime for processing the whole test set and the number of learnable parameters in different models.}
\vskip -0.15in
\label{tab:runtime}
\end{table}

\subsection{Comparison to More Methods and Possible Combinations}\label{sec:exp2}

The methods compared with our DL-Net in Section~\ref{sec:exp1} and~\ref{sec:more_exp} all retain the architecture of the baseline autoencoder. We further considered some other methods that modified the network architecture for utilizing more detailed information of the inpainting and interpolation holes (\ie, the position, spatial size, edge map, etc).

A natural way of utilizing such auxiliary information is to incorporate it into the model input. There exist various ways for achieving the goal~\cite{kim2020texture,wang2018recovering}. We took advantage of spatial feature transforms (SFTs) and inserted the SFT layers~\cite{wang2018recovering} into the encoder of the network to achieve performance gains in handling multiple degradation settings. Specifically, in addition to the degraded image, a binary mask with the same size was also fed into the network. To reduce the memory overhead, we let all the SFT layers share most of their learnable parameters. The training was performed involving the same range of $s$ and $r$ as in Section~\ref{sec:exp1}. The experimental results showed that the SFT empowered model can achieve superior performance in comparison to the autoencoder baseline, and, when combined with our DL-Net, even better performance can be obtained. See Table~\ref{tab:celeba_ii_more} and~\ref{tab:celeba_it_more}, in which the structural similarity index measure (SSIM) is also reported.

We further tested a recent method~\cite{nazeri2019edgeconnect} incorporating edge information in the network input. Since we mostly evaluate the $l_1$ loss, $l_2$ loss, and PSNR for comparing methods, we trained EdgeConnect~\cite{nazeri2019edgeconnect} models using only the reconstruction loss, and we fed the groundtruth edge map extracted by the Canny edge detector into the models, together with each masked image. See also Table~\ref{tab:celeba_ii_more} and~\ref{tab:celeba_it_more} for the experimental performance of the models in image inpainting and image interpolation. We also tested the combination of the model with our DL-Net and it can be seen that significant gains were obtained.
\begin{table}[h!]
\begin{center}\resizebox{0.97\linewidth}{!}{
\begin{tabular}{|C{1.37in}|C{1.25in}|C{1.25in}|}
\hline
Method & \tabincell{c}{$s=20$ \\ $l_1$ loss\,/\,PSNR\,/\,SSIM} &  \tabincell{c}{$s=30$ \\ $l_1$ loss\,/\,PSNR\,/\,SSIM }   \\
\hline\hline
Autoencoder+SFT~\cite{wang2018recovering}          & 0.0276\,/\,31.84\,/\,0.9556 & 0.0489\,/\,26.14\,/\,0.8795  \\   
Combined with DL-Net (ours)                 						 & \textbf{0.0187}\,/\,\textbf{33.24}\,/\,\textbf{0.9678} & \textbf{0.0407}\,/\,\textbf{26.64}\,/\,\textbf{0.8908}    \\ \hline  
\hline\hline
EdgeConnect~\cite{nazeri2019edgeconnect}          & 0.0274\,/\,31.92\,/\,0.9552 & 0.0483\,/\,26.28\,/\,0.8792  \\   
Combined with DL-Net (ours)                 						 & \textbf{0.0187}\,/\,\textbf{33.27}\,/\,\textbf{0.9679} & \textbf{0.0403}\,/\,\textbf{26.76}\,/\,\textbf{0.8914}    \\ \hline  
\end{tabular}}
\end{center}
\caption{Image inpainting with multiple degradations on CelebA: methods combined with DL-Net might lead to even better performance.}
\label{tab:celeba_ii_more}
\vskip -0.18in
\end{table}
\begin{table}[h!]
\begin{center}\resizebox{0.97\linewidth}{!}{
\begin{tabular}{|C{1.37in}|C{1.25in}|C{1.25in}|}
\hline
Method & \tabincell{c}{$r=45\%$ \\ $l_1$ loss\,/\,PSNR\,/\,SSIM } &  \tabincell{c}{$r=75\%$ \\ $l_1$ loss\,/\,PSNR\,/\,SSIM }   \\
\hline\hline
Autoencoder+SFT~\cite{wang2018recovering}          & 0.0348\,/\,31.03\,/\,0.9462 & 0.0596\,/\,26.20\,/\,0.8652  \\   
Combined with DL-Net (ours)                 						 & \textbf{0.0337}\,/\,\textbf{31.13}\,/\,\textbf{0.9468} & \textbf{0.0586}\,/\,\textbf{26.39}\,/\,\textbf{0.8690}    \\ \hline  
\hline\hline
EdgeConnect~\cite{nazeri2019edgeconnect}          & 0.0343\,/\,31.27\,/\,0.9474 & 0.0585\,/\,26.46\,/\,0.8701
  \\   
Combined with DL-Net (ours)                 						 & \textbf{0.0332}\,/\,\textbf{31.37}\,/\,\textbf{0.9478} & \textbf{0.0570}\,/\,\textbf{26.52}\,/\,\textbf{0.8709}    \\ \hline  
\end{tabular}}
\end{center}
\caption{Image interpolation with multiple degradations on CelebA: methods combined with DL-Net might lead to even better performance.}\vskip -0.15in
\label{tab:celeba_it_more}
\end{table}

\begin{table*}[h]
\begin{center}\resizebox{0.95\linewidth}{!}{
\begin{tabular}{|C{1.65in}|C{1.48in}|C{1.48in}|C{1.48in}|C{1.48in}|}
\hline
Method & \tabincell{c}{ $s=10$ \\ $l_1$ loss\,/\,$l_2$ loss\,/\,PSNR } &  \tabincell{c}{$s=20$ \\ $l_1$ loss\,/\,$l_2$ loss\,/\,PSNR } &  \tabincell{c}{$s=30$ \\ $l_1$ loss\,/\,$l_2$ loss\,/\,PSNR }  &  \tabincell{c}{$s=20$,\, centered \\ $l_1$ loss\,/\,$l_2$ loss\,/\,PSNR } \\
\hline\hline
Autoencoder (Default)      & 0.0109\,/\,0.0083\,/\,28.37 & 0.0464\,/\,0.0375\,/\,21.41 & 0.1240\,/\,0.1159\,/\,16.46 & 0.0382\,/\,0.0236\,/\,22.78 \\
Autoencoder (Joint)~\cite{Gao2017}               & 0.0025\,/\,0.0005\,/\,40.33 & 0.0122\,/\,0.0034\,/\,31.94 & 0.0344\,/\,0.0116\,/\,26.26 & 0.0112\,/\,0.0028\,/\,32.90 \\
On-Demand Learning~\cite{Gao2017}                  & 0.0025\,/\,0.0005\,/\,40.28 & 0.0122\,/\,0.0034\,/\,31.94 & 0.0343\,/\,0.0115\,/\,26.26 & 0.0112\,/\,0.0028\,/\,32.87 \\
Multi-Task Learning~\cite{Kendall2018} 		    & 0.0025\,/\,0.0005\,/\,40.36 & 0.0125\,/\,0.0035\,/\,31.77 & 0.0360\,/\,0.0123\,/\,25.94 & 0.0114\,/\,0.0029\,/\,32.74 \\
DL-Net (ours)               						    & \textbf{0.0023}\,/\,\textbf{0.0005}\,/\,\textbf{40.97} & \textbf{0.0116}\,/\,\textbf{0.0031}\,/\,\textbf{32.36} & \textbf{0.0329}\,/\,\textbf{0.0108}\,/\,\textbf{26.59} & \textbf{0.0106}\,/\,\textbf{0.0026}\,/\,\textbf{33.33}   \\ \hline
\end{tabular}}
\end{center}
\caption{Image inpainting with multiple degradations on CelebA: DL-Net compared with competitors in a slightly different training setting and $ I_x + (1-M) \odot \hat{I}_y$.}\vskip -0.2in
\label{tab:celeba_ii_apx}
\end{table*}

\subsection{Single Image Super-Resolution}\label{sec:exp2}

\begin{table*}[t]
\begin{center}\resizebox{0.8\linewidth}{!}{
\begin{tabular}{|C{1.45in}|C{1.25in}|C{1.25in}|C{1.25in}|C{1.in}|}
\hline
Method & \tabincell{c}{ $u=0.0$ \\ PSNR ($\times2$\,/$\times3$\,/$\times4$) } &  \tabincell{c}{$u=1.3$ \\ PSNR ($\times2$\,/$\times3$\,/$\times4$) } &  \tabincell{c}{$u=2.6$ \\ PSNR ($\times2$\,/$\times3$\,/$\times4$) }  &  \#Train. Images  \\
\hline\hline   
Bicubic  			     & 33.66\,/\,30.39\,/\,28.42 & 29.02\,/\,26.57\,/\,24.78 & 26.13\,/\,25.32 \,/\,24.33 & -\\ \hline
VDSR                          & 37.62\,/\,33.89\,/\,31.56 & 30.65\,/\,30.36\,/\,29.88 & 26.39\,/\,26.34\,/\,26.33 &\\
VDSR (DL-Net, ours) & 36.47\,/\,33.28\,/\,31.09 & 36.25\,/\,33.28\,/\,31.17 & 34.97\,/\,32.89\,/\,30.99 &  \\ 
DRRN                          & \textbf{37.82}\,/\,\textbf{34.23}\,/\,31.86 & 30.65\,/\,30.30\,/\,29.83 & 26.39\,/\,26.35\,/\,26.31 & 91+400  \\
DRRN (DL-Net, ours) & 37.32\,/\,33.76\,/\,31.34 & 37.39\,/\,33.83\,/\,31.41 & \textbf{36.76}\,/\,\textbf{33.53}\,/\,31.41 &  \\
IDN                               & 37.73\,/\,34.07\,/\,31.76 & 30.64\,/\,30.32\,/\,29.88 & 26.39\,/\,26.35\,/\,26.33 & \\
IDN (DL-Net, ours)     & 37.07\,/\,33.49\,/\,31.24 & 37.05\,/\,33.58\,/\,31.27 & 36.50\,/\,33.32\,/\,31.14  &  \\ \hline
SRMDNF~\cite{Zhang2018} & 37.79\,/\,34.12\,/\,\textbf{31.96} & \textbf{37.45}\,/\,\textbf{34.16}\,/\,\textbf{31.99} & 34.12\,/\,33.02\,/\,\textbf{31.77}  & 400+800+4744 \\ \hline
\end{tabular}}
\end{center}
\caption{SISR with multiple degradations: models trained using our DL-Net flavored strategy are compared with those trained as suggested. Test results of SRMDNF (which is trained exploiting \emph{$10\times$ more images and substantially more parameters}) are cited from the paper.}
\vskip -0.18in
\label{tab:sisr}
\end{table*}

\begin{table*}[h]
\begin{center}\resizebox{0.95\linewidth}{!}{
\begin{tabular}{|C{1.65in}|C{1.48in}|C{1.48in}|C{1.48in}|C{1.48in}|}
\hline
Method & \tabincell{c}{ $u=0.0$ \\ PSNR ($\times2$\,/$\times3$\,/$\times4$) } &  \tabincell{c}{$u=1.3$ \\ PSNR ($\times2$\,/$\times3$\,/$\times4$) } &  \tabincell{c}{$u=2.6$ \\ PSNR ($\times2$\,/$\times3$\,/$\times4$) }  &  \#Train. Images  \\
\hline\hline
VDSR                          & 33.12\,/\,29.88\,/\,28.15 & 27.93\,/\,27.58\,/\,27.09 & 24.61\,/\,24.58\,/\,24.55 &\\
VDSR (DL-Net, ours) & 32.49\,/\,29.53\,/\,27.86 & 32.28\,/\,29.58\,/\,27.93 & 30.89\,/\,29.30\,/\,27.74 &  \\ 
DRRN                          & 33.39\,/\,30.03\,/\,28.32 & 27.93\,/\,27.53\,/\,27.08 & 24.61\,/\,24.59\,/\,24.54 & 91+400  \\
DRRN (DL-Net, ours) & 33.10\,/\,29.85\,/\,28.06 & 33.18\,/\,29.97\,/\,28.11 & 32.19\,/\,29.77\,/\,28.01 &  \\
IDN                               & 33.28\,/\,29.97\,/\,28.21 & 27.93\,/\,27.54\,/\,27.12 & 24.61\,/\,24.59\,/\,24.56 & \\
IDN (DL-Net, ours)     & 32.83\,/\,29.71\,/\,27.99 & 32.89\,/\,29.73\,/\,28.02 & 32.14\,/\,29.52\,/\,27.92  &  \\ \hline
\end{tabular}}
\end{center}
\caption{SISR with multiple degradations on Set-14: our method compared with the baseline.}\vskip -0.15in
\label{tab:sisr_set14}
\end{table*}

\begin{table*}[h!]
\begin{center}\resizebox{0.95\linewidth}{!}{
\begin{tabular}{|C{1.65in}|C{1.48in}|C{1.48in}|C{1.48in}|C{1.48in}|}
\hline
Method & \tabincell{c}{ $u=0.0$ \\ PSNR ($\times2$\,/$\times3$\,/$\times4$) } &  \tabincell{c}{$u=1.3$ \\ PSNR ($\times2$\,/$\times3$\,/$\times4$) } &  \tabincell{c}{$u=2.6$ \\ PSNR ($\times2$\,/$\times3$\,/$\times4$) }  &  \#Train. Images  \\
\hline\hline
VDSR                          & 31.96\,/\,28.87\,/\,27.34 & 27.66\,/\,27.24\,/\,26.67 & 24.97\,/\,24.94\,/\,24.88 &\\
VDSR (DL-Net, ours) & 31.49\,/\,28.67\,/\,27.20 & 31.44\,/\,28.71\,/\,27.25 & 30.09\,/\,28.57\,/\,27.20 &  \\ 
DRRN                          & 32.13\,/\,29.02\,/\,27.44 & 27.66\,/\,27.21\,/\,26.65 & 24.97\,/\,24.94\,/\,24.87 & 91+400  \\
DRRN (DL-Net, ours) & 31.98\,/\,28.88\,/\,27.32 & 32.10\,/\,28.94\,/\,27.36 & 31.35\,/\,28.88\,/\,27.34 &  \\
IDN                               & 32.05\,/\,28.94\,/\,27.40 & 27.66\,/\,27.22\,/\,26.67 & 24.97\,/\,24.94\,/\,24.88 & \\
IDN (DL-Net, ours)     & 31.83\,/\,28.82\,/\,27.26 & 31.93\,/\,28.89\,/\,27.30 & 31.20\,/\,28.83\,/\,27.30  &  \\ \hline
\end{tabular}}
\end{center}
\caption{SISR with multiple degradations on BSD-500: our method compared with the baseline.} \vskip -0.25in
\label{tab:sisr_bsd}
\end{table*}

\textbf{Network architectures.} We chose popular CNNs as backbones in the SISR experiments, including VDSR (2016)~\cite{Kim2016}, DRRN (2017)~\cite{Tai2017}, and IDN (2018)~\cite{Hui2018}.
These networks are structurally very different such that we are able to know whether our method cooperates well with different off-the-shelf image restoration CNNs. 
See Section~\ref{sec:netarc} for schematic sketches and more details of these backbone networks. 

\textbf{Training and test samples.} In order to be consistent with prior work, we used the famous dataset of 91 images~\cite{Yang2010} for training, along with 400 training images from the Berkeley segmentation dataset (BSD-500)~\cite{Martin2001} which is also widely used.
Our test datasets include Set-5~\cite{Bevilacqua2012}, Set-14~\cite{Zeyde2010_On}, and the official test set of BSD-500 (disjoint from the BSD-500 training set), which consist of 5, 14, and 100 images, respectively.
Following prior arts~\cite{Timofte2013,Dong2014}, we considered only feeding the luminance channel into the networks and simply upscaling the two chrominance channels using bicubic interpolation.
To take full advantage of the self-similarity of natural images, data augmentation including rotation, rescaling, and flipping were also used during training, as in~\cite{Wang2015, Kim2016, Dong2016, Tai2017, Hui2018}.
All training images were downsampled and cropped to get square-shaped training pairs.
For the upscaling factor of 2, 3, and 4, the input\,/\,output image patches for training the SISR models have an uniform width (and also height) of 17\,/\,34, 17\,/\,51, and 17\,/\,68, respectively.
These settings helped us train the reference models successfully.
For training using our DL-Net, we believe larger training patches are more suitable, and hence we adjusted their width to 40\,/\,80, 40\,/\,120, and 40\,/\,160~\cite{Zhang2018}.

\textbf{Training process.} Similarly, we still adopted the ``MSRA'' method~\cite{He2015} to initialize weights in (up)-convolutional layers. 
In order to train the networks reasonably fast, we also took advantage of the ADAM algorithm~\cite{Kingma2015} for stochastic optimization.
As suggested, the base learning rate was always set to 0.001 and the two momentum hyper-parameters were set to be 0.9 and 0.999, respectively.
In correspondence with other research work, we also calculated the average PSNR between ground truth and the generated high-resolution images on the luminance channel (in the YCrCb color space).
Since a specific upscaling layer is introduced for each scale, one model should be trained each for the $\times2$, $\times3$, and $\times4$ scenarios.
We cut the learning rate by $10\times$ every 80 epochs, such that the model performance was finally saturated~\cite{Kim2016}.
After training for 300 epochs, our VDSR and DRRN $\times 3$ models achieved average PSNRs of 33.89dB and 34.23dB on Set-5, respectively.

\textbf{Main results.} The above training process follows that of existing deep SISR CNNs in which all input low-resolution images were assumed to be self-degenerated directly through bicubic interpolation using high-resolution images. 
As have been discussed, such an assumption should fail on real-world applications where blur (or other types of degradations) also exist.
To verify this (just as with the image inpainting task), we first evaluated the obtained reference models on low-resolution images downscaled from (probably) blurry images.
We chose the same blur kernels as in~\cite{Zhang2018} whose width $u$ vary in $[0,3]$.
Our results (in Table~\ref{tab:sisr}) demonstrate the performance of state-of-the-art SISR models diminishes a lot when some subtle distortion is introduced. 

Fortunately, the proposed DL-Net strategy helps to resolve this problem and improve the SISR performance to a remarkable extent. 
As demonstrated in Table~\ref{tab:sisr}, our DL-Net models with VDSR, DRRN, and IDN as the backbone architectures all achieve prominent results for both $u=1.3$ and $2.6$, while in an ideal setting (i.e., $u=0$) their PSNR indices drop only a little bit.
We also compared our method with a very recent method dedicated to addressing the fixation problem for SISR~\cite{Zhang2018}.
Though trained exploiting $10\times$ more images and substantially more parameters ($\sim1.49$M, while our DRRN: $\sim0.36$M), its performance \emph{still diminishes} in the less challenging $\times2$ and $\times 3$ scenarios with $u=2.6$.
Compared with SRMDNF and the references, our DL-Net shows more stable performance across all degradation settings. 
In addition, though our work mostly focus on \emph{image} restoration, it is possible to adopt our DL-Net to \emph{video} processing, \eg, video super-resolution. We tried using our DL-Net on the basis of VESPCN~\cite{caballero2017real}, and we indeed observed performance gain in video super-resolution. Specifically, for $\times2$ super-resolution, the VESPCN baseline achieved $33.76$dB, $28.46$dB, and $25.61$dB, for $u=0.0$, $1.3$, and $2.6$ on videos degraded from \url{https://www.harmonicinc.com/4k-demo-footage-download}, respectively, while using our DL-Net led to $33.41$dB, $28.75$dB, and $25.95$dB in the sense of PSNR. We did not perform large-scale experiments for this since it is out of the scope of this paper.

Experimental results on Set-14 and BSD-500 are given in Table~\ref{tab:sisr_set14} and~\ref{tab:sisr_bsd}.
Obviously, better results can be obtained using our DL-Net, when additional blurring is inevitable.
It is also worth noting that, with our method, similar PSNR can be obtained under various levels of SISR degradations. 
We don't have qualitative results of SRMDNF on the two popular datasets, so only the baseline is taken for comparison. 

\begin{figure*}[h!]
\captionsetup[subfigure]{labelformat=empty}
\begin{center}
\subfloat[Input\hspace{3em} $u=2.6$ ]{
\includegraphics[width=0.07\textwidth]{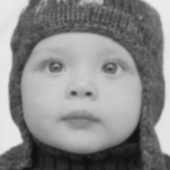}}\hskip 18pt
\subfloat[$u=0.0$, DRRN]{
\includegraphics[width=0.21\textwidth]{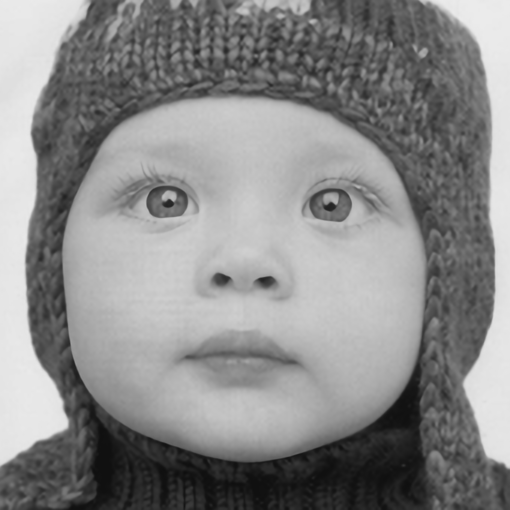}}\hskip 18pt
\subfloat[$u=1.3$, DRRN]{
\includegraphics[width=0.21\textwidth]{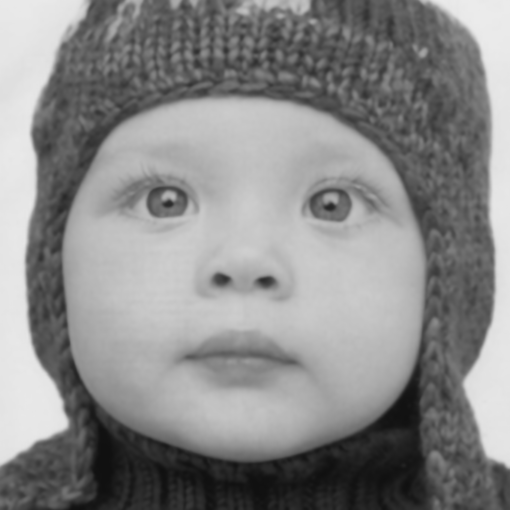}}\hskip 18pt
\subfloat[$u=2.6$, DRRN]{
\includegraphics[width=0.21\textwidth]{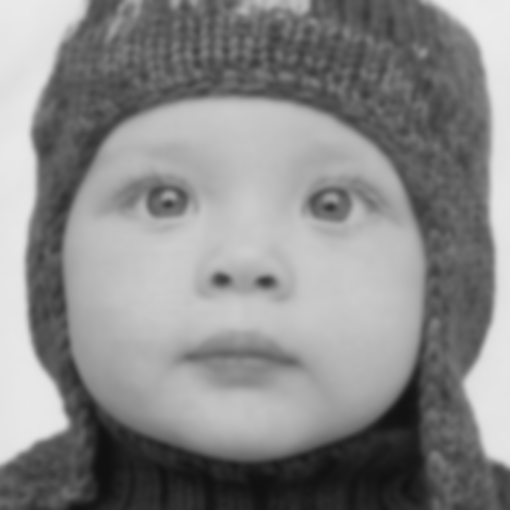}}\hskip 18pt \\

\hskip 54pt
\subfloat[$u=0.0$, DRRN (DL-Net)]{
\includegraphics[width=0.21\textwidth]{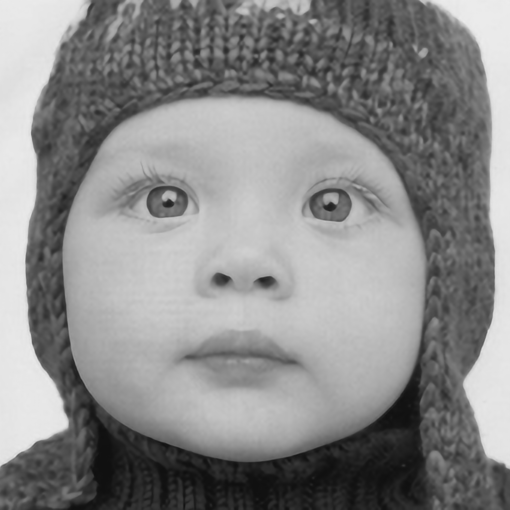}}\hskip 18pt
\subfloat[$u=1.3$, DRRN (DL-Net)]{
\includegraphics[width=0.21\textwidth]{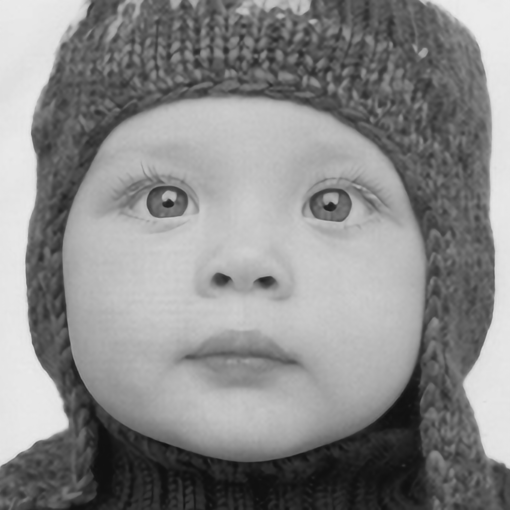}}\hskip 18pt
\subfloat[$u=2.6$, DRRN (DL-Net)]{
\includegraphics[width=0.21\textwidth]{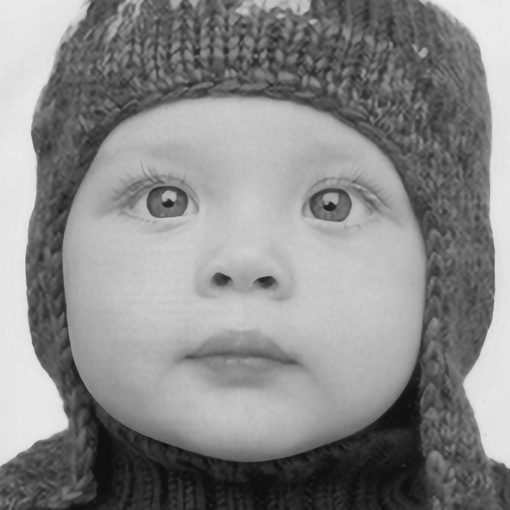}}\hskip 18pt \\ \vskip 1em

\subfloat[Input\hspace{3em} $u=2.6$]{
\includegraphics[width=0.07\textwidth]{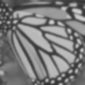}}\hskip 18pt
\subfloat[$u=0.0$, DRRN]{
\includegraphics[width=0.21\textwidth]{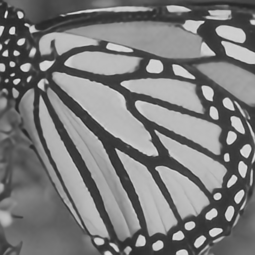}}\hskip 18pt
\subfloat[$u=1.3$, DRRN]{
\includegraphics[width=0.21\textwidth]{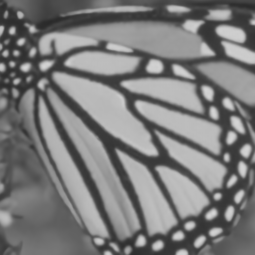}}\hskip 18pt
\subfloat[$u=2.6$, DRRN]{
\includegraphics[width=0.21\textwidth]{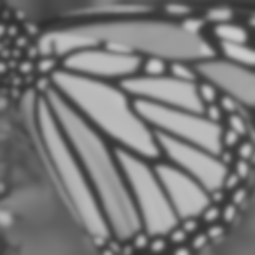}}\hskip 18pt \\

\hskip 54pt
\subfloat[$u=0.0$, DRRN (DL-Net)]{
\includegraphics[width=0.21\textwidth]{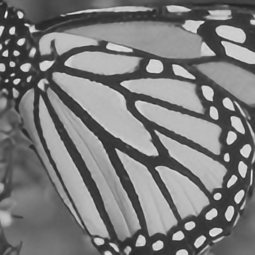}}\hskip 18pt
\subfloat[$u=1.3$, DRRN (DL-Net)]{
\includegraphics[width=0.21\textwidth]{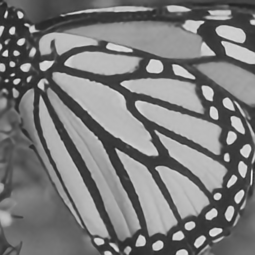}}\hskip 18pt
\subfloat[$u=2.6$, DRRN (DL-Net)]{
\includegraphics[width=0.21\textwidth]{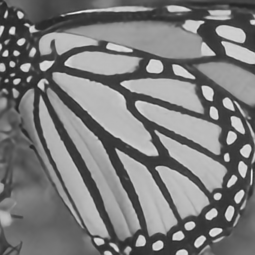}}\hskip 18pt 

\caption{SISR results with the original DRRN and the retrained model powered by our method.} 
\label{fig:6}
\end{center}
\vskip -0.13in
\end{figure*}

\textbf{Qualitative Results.} We also provide qualitative results for SISR under multiple degradation settings. See Figure~\ref{fig:6}, in which we illustrate the luminance channel only to enable comparison between direct outputs of different network models.
It can be seen that our obtained models show perceptually similar results in different degradation settings while the performance of the original DRRN diminishes significantly under $u=1.3$ and $u=2.6$.

\subsection{Network Architectures}\label{sec:netarc}
In this subsection, we elaborate on the network architectures introduced in the main body of our paper. 

First, it is our inpainting and interpolation autoencoders.
The encoder and decoder parts each consists of four (up)-convolutional layers, followed by batch normalization~\cite{Ioffe2015} and nonlinear activations.
A channel-wise fully-connected layer that propagates information only within feature maps is used to concatenate the two parts.
While leaky ReLUs were used in the encoder, ReLUs were directly adopted in the decoder for nonlinearity, just like the context encoder network~\cite{Pathak2016}.

To be more specific, its first half (i.e., the encoder part) is with: conv1 (kernel: $[4\times4,\ 64]$, stride: 2), conv2 (kernel: $[4\times4,\ 128]$, stride: 2), conv3 (kernel: $[4\times4,\ 256]$, stride: 2), and conv2 (kernel: $[4\times4,\ 512]$, stride: 2). 
The channel-wise fully-connected layer consists of 512 $16\times 16$ filters.
The second half (i.e., the decoder) consists of four convolutional layers with stride 1/2 (or equivalently deconvolutions with stride 2), being structurally symmetric to the encoder. 
Batch normalization was used before (leaky) ReLU, as prescribed in the paper~\cite{Gao2017, Pathak2016}.
A graphical demonstration of the network architecture can be found in~\cite{Gao2017}.

\begin{figure*}[t]
\begin{center}
\subfloat[]{\label{fig:5a}
\includegraphics[width=0.24\textwidth]{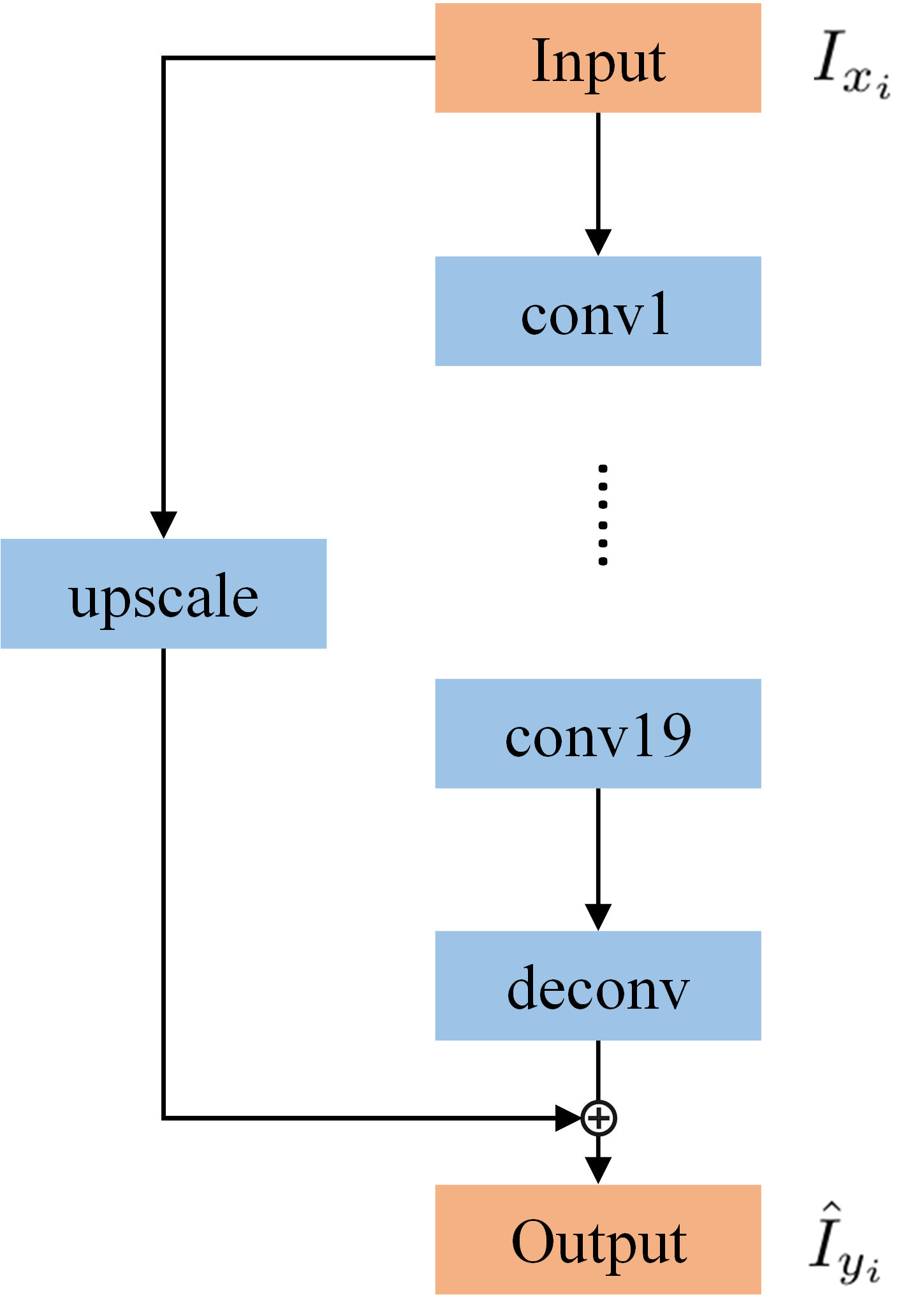}}\hskip 18pt
\subfloat[]{\label{fig:5b}
\includegraphics[width=0.24\textwidth]{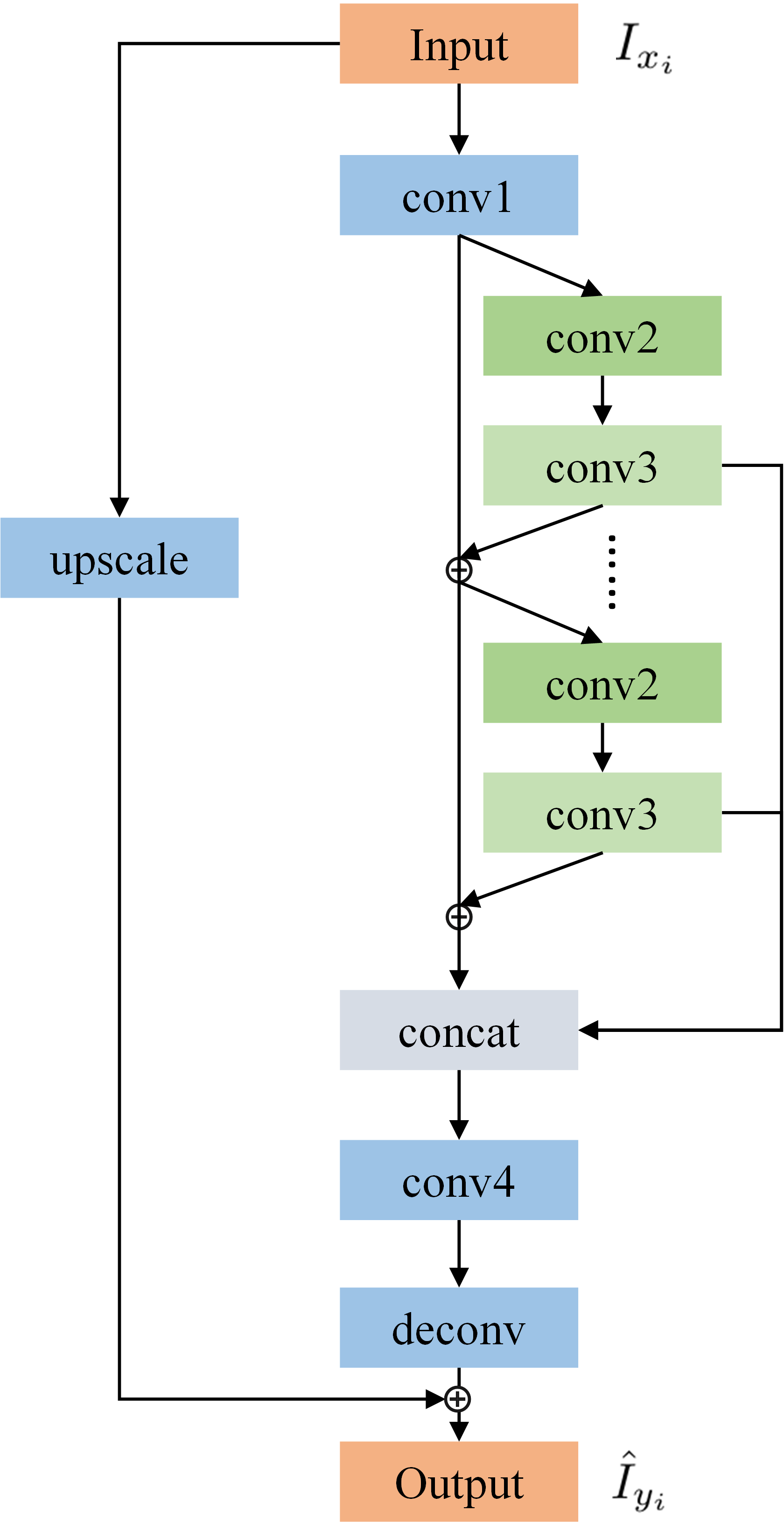}}\hskip 18pt
\subfloat[]{\label{fig:5c}
\includegraphics[width=0.24\textwidth]{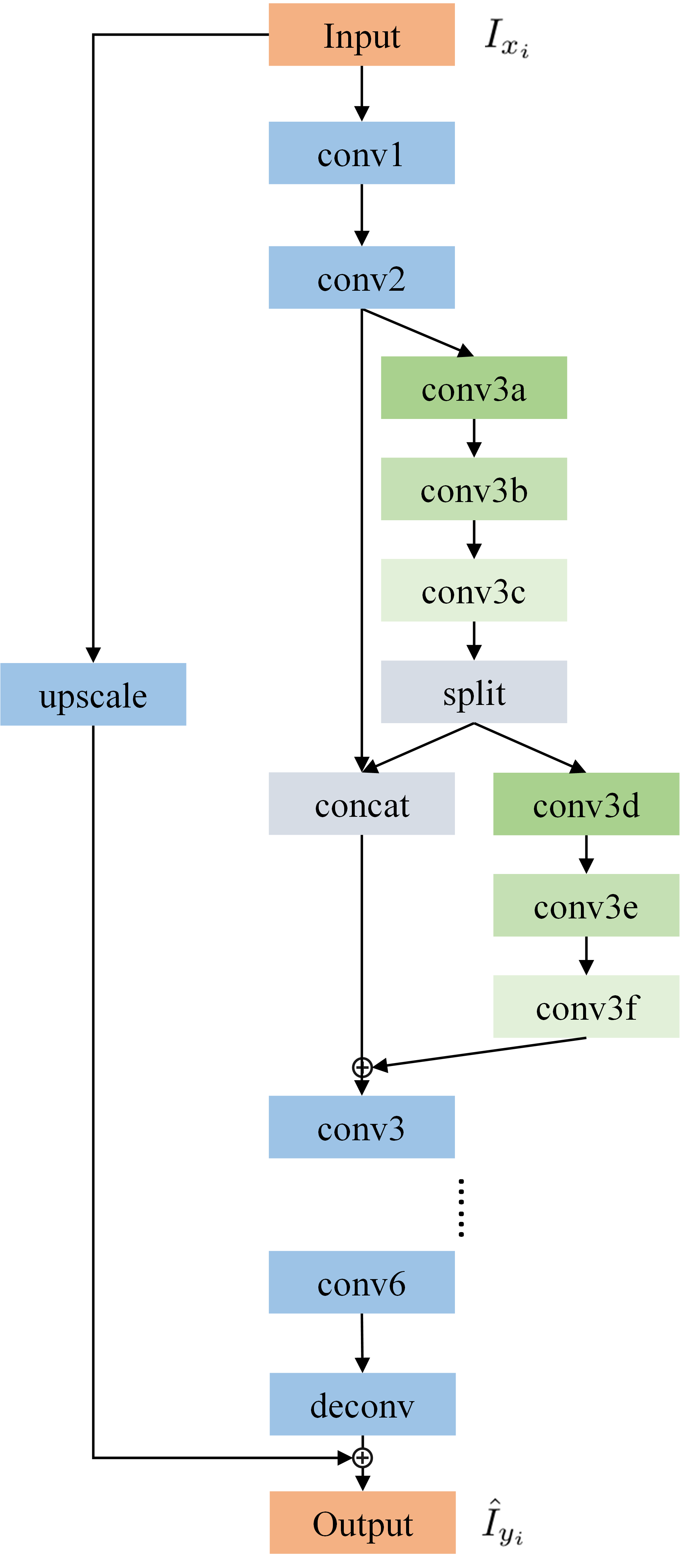}}
\caption{The schematic sketches for our (a) VDSR, (b) DRRN, and (c) IDN.}
\label{fig:5}
\end{center}
\vskip -0.15in
\end{figure*}

For the three SISR networks adopted in our paper, we provide schematic sketches for their architectures in Figure~\ref{fig:5}.
VDSR is a 20-layer CNN that takes advantage of residual learning~\cite{He2016} and also establishes a (global) skip connection from its input to the output.
DRRN aims to further deepen the network and it adopts residual learning in both global and local manners. 
To make the computation more efficient, we adapt the networks to manipulate image features on a low-resolution level~\cite{Dong2016} by modifying the global skip connection to a bicubic upscaling layer as in IDN~\cite{Hui2018}.
The width of convolutional layers in VDSR and DRRN are uniformly set to be 64 and 128, as suggested in the papers.
Specifically, for DRRN, we adopt ``DRRN\_B1\_U9'' and further introduce a concat layer that merges sequential results from the recursive block to boost its performance. 
The concatenation is performed every three recursive steps, which means useful knowledge can be distilled from $128\times 3=384$ feature maps in total.
The conv4 layer of our DRRN conducts $1\times 1$ convolutions and also outputs 64 feature maps. 
Also, inspired by IDN~\cite{Hui2018}, we use leaky ReLU with a negative slop 0.05 instead of the original ReLU in DRRN to prevent the so-called ``dying ReLU'' problem.
For IDN, our configurations mostly follow those suggested in the paper~\cite{Hui2018}.
Its final performance with $u=0$ might be a little bit lower than that reported in its paper~\cite{Hui2018}, partially because we only use the $l_2$ loss for simplicity of training.

\section{More Discussions about Related Work}\label{sec:mdc}

Our DL-Net is related with some prior work and we have briefly introduced them in Section 2. 
More detailed discussions will be given in this section. 
We believe that, after careful explanations in the prequel of the paper, the connections between our method and other work should be more understandable to the readers.

It has long been known that deep CNNs extract contextual information from ground-truth high-quality images. 
Natural image priors were introduced to encourage smooth textures.
However, only until very recently had we been aware of the prior knowledge brought in with the deep architecture itself.
Considering implicit priors captured by the network, Ulyanov \etal\cite{Ulyanov2018} proposed to maximize task-dependent likelihood for pursuing decent image restoration performance.
Our DL-Net schematically suggests output images that being able to reproduce the corresponding inputs (i.e., maximize the likelihood as well), thus it seems similar to Ulyanov et al.'s deep image prior~\cite{Ulyanov2018}.
Yet, the superiority of our method also rests on rich supervision from many high-quality training images and insightful knowledge extracted from the degradations. 
Benefit from external data and the degradation information, our DL-Net outperforms some other supervised methods and its computational cost is relatively low.
By contrast, though the deep image prior also applies to image restoration with multiple degradations, its performance is only comparable with the supervised state-of-the-art, and it requires thousands of iterations to run on a single test image.   

Our method is also related with AffGAN~\cite{Sonderby2017} in which the amortized MAP inference was explored for SISR and a projection layer was introduced to guarantee its likelihood-based constraints being explicitly satisfied.
Such a projection layer advocates outcomes fulfilling its implicit assumptions and low likelihood loss is naturally obtained.
However, AffGAN mostly focused on simple SISR problems whose given inputs were down-sampled through only a presumed bicubic interpolation, and we stress that AffGAN does not apply to our task where multiple blurring levels exist. This is mainly because a single or even several $\prod^A_x$ operations cannot guarantee the constraints anymore in our setting with ($u\in[0,3]$). 

\section{Conclusions}\label{sec:con}
 
While impressive results have been gained, state-of-the-art image restoration networks were usually trained with self-degenerated images obtained in a very restricted degradation setting, often also in a particular level.  
Such limitation should hinder restoration networks from being applied to some real-world applications, and there is as of yet no general solution to address the limitation.
We have proposed DL-Net in this paper, towards generalizing existing networks to succeed over a spectrum of degradation levels with their training objective and core architectures reused.
The pivotal of our method is to assist a subnet $h$ for disentangling the effects of possible degradations and for minimizing $\mathcal L$.
Experimental results on image inpainting, image interpolation, and SISR have verified the effectiveness of our method.
Future work shall include explorations on more degradation types.

\section*{Acknowledgements}
This work is funded by the Natural Science Foundation of China (NSFC) and the German Research Foundation (DFG) in the project of Cross-modal Learning (NSFC 62061136001/ DFG TRR-169 and NSFC 61876095).

\bibliographystyle{IEEEtran}
\bibliography{ref}

\begin{IEEEbiography}[{\includegraphics[width=1in,height=1.25in,clip,keepaspectratio]{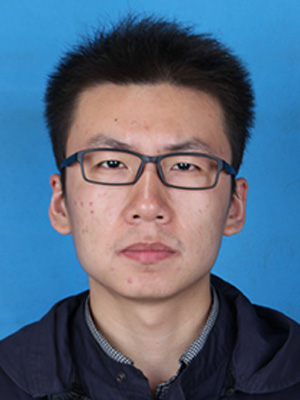}}]{Yiwen Guo} received the B.E. degree from Wuhan University, Wuhan, China, in 2011, and the Ph.D. degree from Tsinghua University, Beijing, China in 2016. He is a research scientist at Bytedance AI Lab, Beijing. Prior to this, he was a staff research scientist at Intel Labs China. His current research interests include computer vision, pattern recognition, and machine learning.
\end{IEEEbiography}

\begin{IEEEbiography}[{\includegraphics[width=1in,height=1.25in,clip,keepaspectratio]{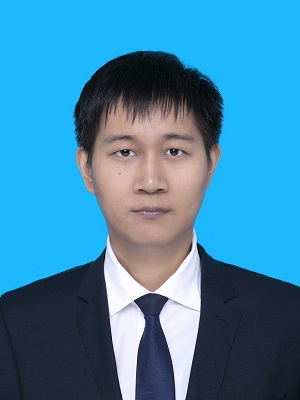}}]{Ming Lu} Ming Lu received the PhD degree in Information and Communication Engineering from Tsinghua University, Beijing, China, in 2019. He is currently a researcher at Intel Labs China. His research interests include computer vision and computer graphics. He is particularly interested in classification/detection, 3D face/body, and image restoration/synthesis.\end{IEEEbiography}

\begin{IEEEbiography}[{\includegraphics[width=1in,height=1.25in,clip,keepaspectratio]{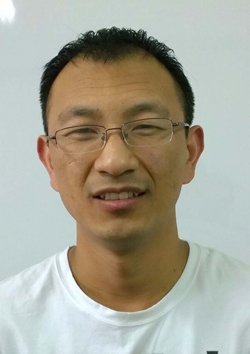}}]{Wangmeng Zuo} (M'09, SM'14) received the Ph.D. degree in computer application technology from the Harbin Institute of Technology, China, in 2007. From 2004 to 2006, he was a Research Assistant with the Department of Computing, The Hong Kong Polytechnic University. From 2009 to 2010, he was a Visiting Professor with Microsoft Research Asia. He is currently a Professor with the School of Computer Science and Technology, Harbin Institute of Technology. He has published over 90 papers in top-tier academic journals and conferences. His current research interests include image enhancement and restoration, image generation and editing, visual tracking, object detection, and image classification. He has served as a Tutorial Organizer in ECCV 2016, an Associate Editor of the IET Biometrics, and the Guest Editor of Neurocomputing, Pattern Recognition, IEEE Transactions on Circuits and Systems for Video Technology, and IEEE Transactions on Neural Networks and Learning Systems.
\end{IEEEbiography}

\begin{IEEEbiography}[{\includegraphics[width=1in,height=1.25in,clip,keepaspectratio]{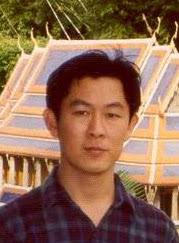}}]{Changshui Zhang} received the B.E. degree in mathematics from Peking University, Beijing, China, in 1986, and the M.S. and Ph.D. degrees in control science and engineering from Tsinghua University, Beijing, in 1989 and 1992, respectively. In 1992, he joined the Department of Automation, Tsinghua University, where he is currently a professor. His research interests include pattern recognition and machine learning. He is a Fellow member of the IEEE.
\end{IEEEbiography}

\begin{IEEEbiography}[{\includegraphics[width=1in,height=1.25in,clip,keepaspectratio]{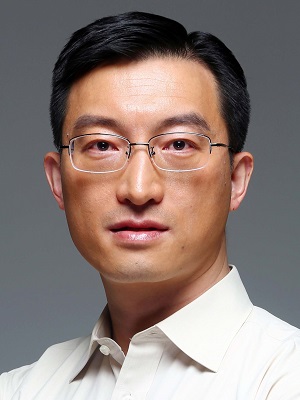}}]{Yurong Chen} received the B.S. and Ph.D. degrees from Tsinghua University, Beijing, China, in 1998 and 2002, respectively. He joined Intel in 2004 after completing the postdoctoral research in the Institute of Software, CAS, where he is currently a Principal Research Scientist and Director of Cognitive Computing Lab at Intel Labs China, responsible for leading visual cognition and machine learning research for Intel platforms. He received one “Intel China Award” and 3 Intel Labs Academic Awards – “Gordy Awards” for delivering leading visual analytics and understanding technologies to impact Intel platforms/solutions. He has published over 60 papers and holds over 50 issued/pending patents. 
\end{IEEEbiography}

\end{document}


\title{Deep Likelihood Network for Image Restoration with Multiple Degradation Levels **Appendices**}

\author{Yiwen Guo, Ming Lu, Wangmeng Zuo, Changshui Zhang,~\IEEEmembership{Fellow,~IEEE}, and Yurong Chen
\thanks{Y. Guo is with Bytedance AI Lab. E-mail: guoyiwen.ai@bytedance.com.}
\thanks{W. Zuo is with the School of Computer Science and Technology, Harbin Institute of Technology, Harbin 150001, China. E-mail: cswmzuo@gmail.com.}
\thanks{M. Lu and Y. Chen are with Intel Labs China, Beijing 100190, China. E-mail: ming1.lu@intel.com, yurong.chen@intel.com.}
\thanks{C. Zhang is with the Department of Automation, State Key Lab of Intelligence Technologies and Systems, Tsinghua National Laboratory for Information Science and Technology, Tsinghua University, Beijing 100084, China. E-mail: zcs@mail.tsinghua.edu.cn}
\thanks{Manuscript received Nov. 12, 2019.}}

\markboth{SUBMISSION TO IEEE TRANSACTIONS ON IMAGE PROCESSING}{}


\maketitle

%

\IEEEpeerreviewmaketitle

\appendices
%

\section{More About Image Inpainting}\label{sec:moreinp}

We were aware that there might exist different choices for the formulation of $\mathcal L_{rec}$ when training inpainting networks.
In addition to the one derived in Eq.~(5), researchers also propose to enlarge the inpainting blocks by 7 pixels and penalize $10\times$ more on the boundary regions for encouraging perceptual consistency~\cite{Pathak2016}.
Here we also report experimental results on CelebA in such setting.
As shown in Table~\ref{tab:celeba_ii_apx}, the superiority of our method holds consistently.

\begin{table*}[h]\addtocounter{table}{+9}
\begin{center}\resizebox{0.95\linewidth}{!}{
\begin{tabular}{|C{1.65in}|C{1.48in}|C{1.48in}|C{1.48in}|C{1.48in}|}
\hline
Method & \tabincell{c}{ $s=10$ \\ $l_1$ loss\,/\,$l_2$ loss\,/\,PSNR } &  \tabincell{c}{$s=20$ \\ $l_1$ loss\,/\,$l_2$ loss\,/\,PSNR } &  \tabincell{c}{$s=30$ \\ $l_1$ loss\,/\,$l_2$ loss\,/\,PSNR }  &  \tabincell{c}{$s=20$,\, centered \\ $l_1$ loss\,/\,$l_2$ loss\,/\,PSNR } \\
\hline\hline
Autoencoder (Default)      & 0.0109\,/\,0.0083\,/\,28.37 & 0.0464\,/\,0.0375\,/\,21.41 & 0.1240\,/\,0.1159\,/\,16.46 & 0.0382\,/\,0.0236\,/\,22.78 \\
Autoencoder (Joint)~\cite{Gao2017}               & 0.0025\,/\,0.0005\,/\,40.33 & 0.0122\,/\,0.0034\,/\,31.94 & 0.0344\,/\,0.0116\,/\,26.26 & 0.0112\,/\,0.0028\,/\,32.90 \\
On-Demand Learning~\cite{Gao2017}                  & 0.0025\,/\,0.0005\,/\,40.28 & 0.0122\,/\,0.0034\,/\,31.94 & 0.0343\,/\,0.0115\,/\,26.26 & 0.0112\,/\,0.0028\,/\,32.87 \\
Multi-Tasks Learning~\cite{Kendall2018} 		    & 0.0025\,/\,0.0005\,/\,40.36 & 0.0125\,/\,0.0035\,/\,31.77 & 0.0360\,/\,0.0123\,/\,25.94 & 0.0114\,/\,0.0029\,/\,32.74 \\
DL-Net (ours)               						    & \textbf{0.0023}\,/\,\textbf{0.0005}\,/\,\textbf{40.97} & \textbf{0.0116}\,/\,\textbf{0.0031}\,/\,\textbf{32.36} & \textbf{0.0329}\,/\,\textbf{0.0108}\,/\,\textbf{26.59} & \textbf{0.0106}\,/\,\textbf{0.0026}\,/\,\textbf{33.33}   \\ \hline
\end{tabular}}
\end{center}
\caption{Image interpolation with multiple degradations on CelebA: DL-Net compared with competitors in a slightly different training setting and $ I_x + (1-M) \odot \hat{I}_y$.}\vskip -0.2in
\label{tab:celeba_ii_apx}
\end{table*}

\section{SISR on Set-14 and BSD-500}\label{sec:sisrmore}

We report SISR results on Set-14 and BSD-500 (in which only the 100 test images are used in the evaluation) in this section.
All training and test policies are kept the same as on Set-5, and the results are summarized in Table~\ref{tab:sisr_set14} and~\ref{tab:sisr_bsd}.
Obviously, better results can be obtained using our DL-Net strategy, when some additional blurring is inevitable.
It is also worthwhile noting that, with our method, similar PSNR performance can be obtained under various level of SISR degradations. 
We don't have qualitative results of SRMDNF on the two popular datasets, so  only the baseline is taken for comparison.

\begin{table*}[h]
\begin{center}\resizebox{0.95\linewidth}{!}{
\begin{tabular}{|C{1.65in}|C{1.48in}|C{1.48in}|C{1.48in}|C{1.48in}|}
\hline
Method & \tabincell{c}{ $u=0.0$ \\ PSNR ($\times2$\,/$\times3$\,/$\times4$) } &  \tabincell{c}{$u=1.3$ \\ PSNR ($\times2$\,/$\times3$\,/$\times4$) } &  \tabincell{c}{$u=2.6$ \\ PSNR ($\times2$\,/$\times3$\,/$\times4$) }  &  \#Train. Images  \\
\hline\hline
VDSR                          & 33.12\,/\,29.88\,/\,28.15 & 27.93\,/\,27.58\,/\,27.09 & 24.61\,/\,24.58\,/\,24.55 &\\
VDSR (DL-Net, ours) & 32.49\,/\,29.53\,/\,27.86 & 32.28\,/\,29.58\,/\,27.93 & 30.89\,/\,29.30\,/\,27.74 &  \\ 
DRRN                          & 33.39\,/\,30.03\,/\,28.32 & 27.93\,/\,27.53\,/\,27.08 & 24.61\,/\,24.59\,/\,24.54 & 91+400  \\
DRRN (DL-Net, ours) & 33.10\,/\,29.85\,/\,28.06 & 33.18\,/\,29.97\,/\,28.11 & 32.19\,/\,29.77\,/\,28.01 &  \\
IDN                               & 33.28\,/\,29.97\,/\,28.21 & 27.93\,/\,27.54\,/\,27.12 & 24.61\,/\,24.59\,/\,24.56 & \\
IDN (DL-Net, ours)     & 32.83\,/\,29.71\,/\,27.99 & 32.89\,/\,29.73\,/\,28.02 & 32.14\,/\,29.52\,/\,27.92  &  \\ \hline
\end{tabular}}
\end{center}
\caption{SISR with multiple degradations on Set-14: our method compared with the baseline.}\vskip -0.15in
\label{tab:sisr_set14}
\end{table*}

\begin{table*}[h!]
\begin{center}\resizebox{0.95\linewidth}{!}{
\begin{tabular}{|C{1.65in}|C{1.48in}|C{1.48in}|C{1.48in}|C{1.48in}|}
\hline
Method & \tabincell{c}{ $u=0.0$ \\ PSNR ($\times2$\,/$\times3$\,/$\times4$) } &  \tabincell{c}{$u=1.3$ \\ PSNR ($\times2$\,/$\times3$\,/$\times4$) } &  \tabincell{c}{$u=2.6$ \\ PSNR ($\times2$\,/$\times3$\,/$\times4$) }  &  \#Train. Images  \\
\hline\hline
VDSR                          & 31.96\,/\,28.87\,/\,27.34 & 27.66\,/\,27.24\,/\,26.67 & 24.97\,/\,24.94\,/\,24.88 &\\
VDSR (DL-Net, ours) & 31.49\,/\,28.67\,/\,27.20 & 31.44\,/\,28.71\,/\,27.25 & 30.09\,/\,28.57\,/\,27.20 &  \\ 
DRRN                          & 32.13\,/\,29.02\,/\,27.44 & 27.66\,/\,27.21\,/\,26.65 & 24.97\,/\,24.94\,/\,24.87 & 91+400  \\
DRRN (DL-Net, ours) & 31.98\,/\,28.88\,/\,27.32 & 32.10\,/\,28.94\,/\,27.36 & 31.35\,/\,28.88\,/\,27.34 &  \\
IDN                               & 32.05\,/\,28.94\,/\,27.40 & 27.66\,/\,27.22\,/\,26.67 & 24.97\,/\,24.94\,/\,24.88 & \\
IDN (DL-Net, ours)     & 31.83\,/\,28.82\,/\,27.26 & 31.93\,/\,28.89\,/\,27.30 & 31.20\,/\,28.83\,/\,27.30  &  \\ \hline
\end{tabular}}
\end{center}
\caption{SISR with multiple degradations on BSD-500: our method compared with the baseline.} \vskip -0.25in
\label{tab:sisr_bsd}
\end{table*}

\section{Network Architectures}\label{sec:netarc}

In this section, we elaborate on the network architectures introduced in the main body of our paper. 
First it is our inpainting and interpolation autoencoder.
The encoder and decoder parts are each comprised of four (up)-convolutional layers, all followed by batch normalizations~\cite{Ioffe2015} and nonlinear activations.
A channel-vise fully-connected layer that propagates information only within feature maps is used to concatenate the two parts.
While leaky ReLUs are used in the encoder, ReLUs are directly adopted in the decoder for nonlinearity, just like the context encoder network~\cite{Pathak2016}.

To be more specific, its first half (i.e., the encoder part) is with: conv1 (kernel: $[4\times4,\ 64]$, stride: 2), conv2 (kernel: $[4\times4,\ 128]$, stride: 2), conv3 (kernel: $[4\times4,\ 256]$, stride: 2) and conv2 (kernel: $[4\times4,\ 512]$, stride: 2). 
The channel-wise fully-connected layer consists of 512 $16\times 16$ filters.
The second half (i.e., the decoder) consists of four convolutional layers with stride 1/2 (or equivalently deconvolutions with stride 2), being structurally symmetric to the encoder. 
Batch normalization is used before (leaky) ReLU, as prescribed in the paper~\cite{Gao2017, Pathak2016}.
A graphical demonstration of the network architecture can be found in~\cite{Gao2017}.

Fo the three SISR networks adopted in our paper, we provide schematic sketches for their architectures in Figure~\ref{fig:5}.
VDSR is a 20-layer CNN that takes advantage of residual learning~\cite{He2016} and for the first time builds a (global) skip connection from its input to the output.
DRRN aims to further deepen the network and adopts residual learning in both global and local manners. 
To make the computation more efficient, we adapt them to manipulate image features on a low-resolution level~\cite{Dong2016} by modifying the global skip connection to a bicubic upscaling layer as in IDN~\cite{Hui2018}.
The width of convolutional layers in VDSR and DRRN are uniformly set to be 64 and 128, as suggested in the papers.
Specifically, for DRRN, we adopt ``DRRN\_B1\_U9'' and further introduce a concat layer that merges sequential results from the recursive block to boost its performance. 
The concatenation is performed every three recursive steps, which means useful knowledge can be distilled from $128\times 3=384$ feature maps in total.
The conv4 layer of our DRRN conducts $1\times 1$ convolutions and also outputs 64 feature maps. 
Also, inspired by IDN~\cite{Hui2018}, we use leaky ReLU with a negative slop 0.05 instead of the original ReLU in DRRN to prevent the so-called ``dying ReLU'' problem.
For IDN, our configurations mostly follow those suggested in the paper~\cite{Hui2018}.
Its final performance with $u=0$ might be a little bit lower than that reported in its paper~\cite{Hui2018}, partially because we only use the $l_2$ loss for simplicity of training.

\section{More Discussions and Comparisons}\label{sec:mdc}

Our DL-Net is related with some previous works and we have briefly introduced them in Section 2. 
We will give some more detailed discussions in this section.
It has long been known that deep CNNs extract contextual information from ground-truth high-quality images. 
Natural image priors are introduced to encourage smooth textures.
However, only until recently have we been aware of the prior knowledge brought in with the deep architecture itself~\cite{Ulyanov2018}.
Considering implicit priors captured by the network, Ulyanov et al., propose to directly minimize is a task-dependent likelihood for pursuing decent image restoration performance.

Our DL-Net schematically suggests outputs that being able to reproduce the corresponding inputs (i.e., minimize the likelihood), which might seem similar to Ulyanov et al.'s deep image prior. 
In fact, the superiority of our method also rests on rich supervision from numerous real images and some insightful knowledge extracted from the degradations. 
Benefit from external data and the degradation information, our DL-Net outperforms some other supervised methods and its computational complexity is relatively low.
Although Ulyanov et al.'s method also applies to restoration with multiple degradations, their performance is only comparable with the supervised state-of-the-art, and it requires thousands of iterations to run on a single test image.   

Our method is also related with AffGAN~\cite{Sonderby2017} in which the amortized MAP inference is explored for SISR and a projection layer is introduced to guarantee its likelihood-based constraints being explicitly satisfied.
Such a projection layer advocates outcomes strictly fulfilling its implicit assumptions and low likelihood loss is naturally obtained.
However, AffGAN focuses on simple SISR problems whose given inputs are down-sampled through only a presumed bicubic interpolation.
We stress that it does not apply to our task where multiple blurring levels exist, mainly because a single or even several $\prod^A_x$ operations cannot guarantee the constraints anymore in our setting with ($u\in[0,3]$). 


\begin{figure*}[t]\addtocounter{figure}{+4}
\begin{center}
\subfloat[]{\label{fig:5a}
\includegraphics[width=0.24\textwidth]{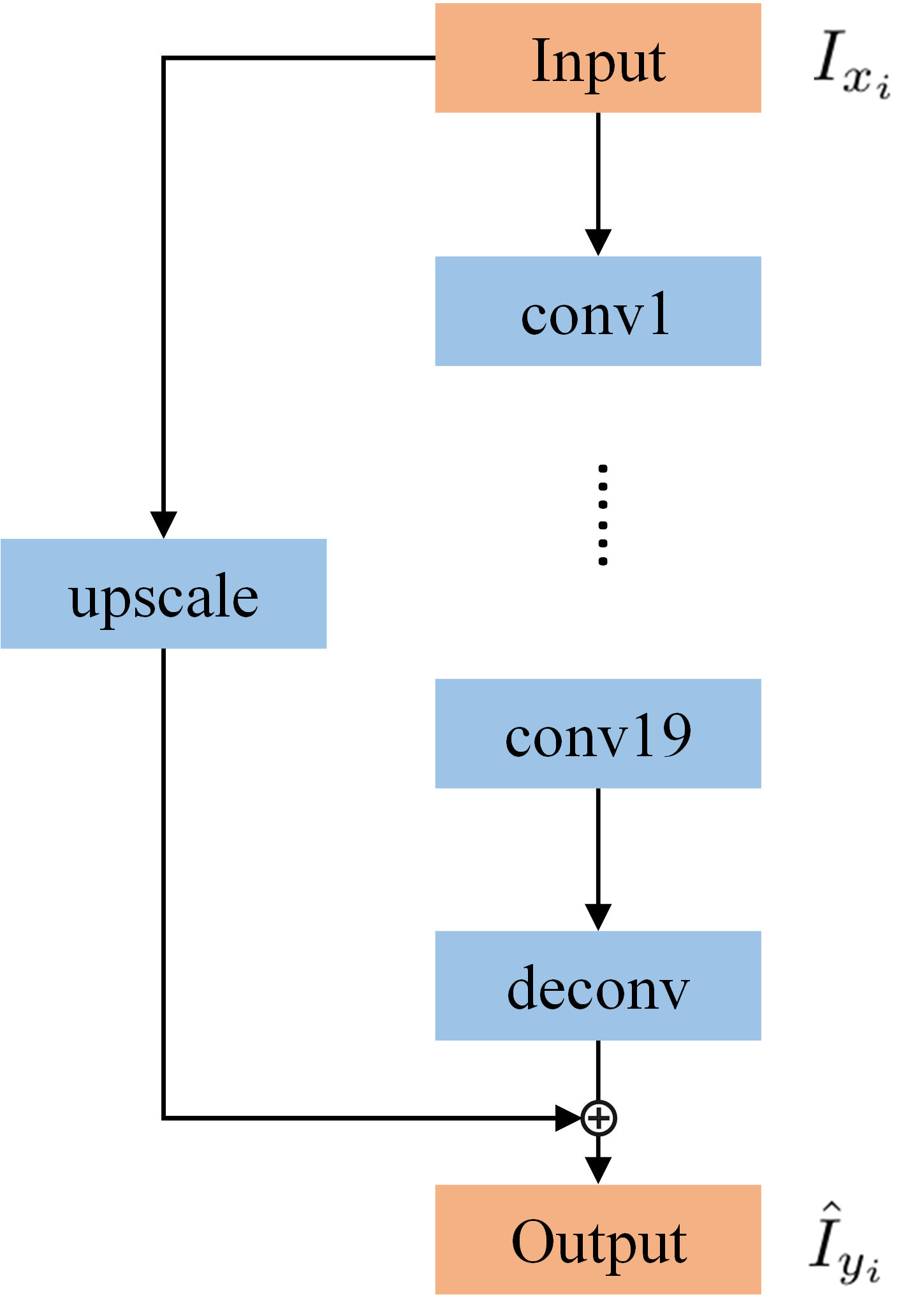}}\hskip 18pt
\subfloat[]{\label{fig:5b}
\includegraphics[width=0.24\textwidth]{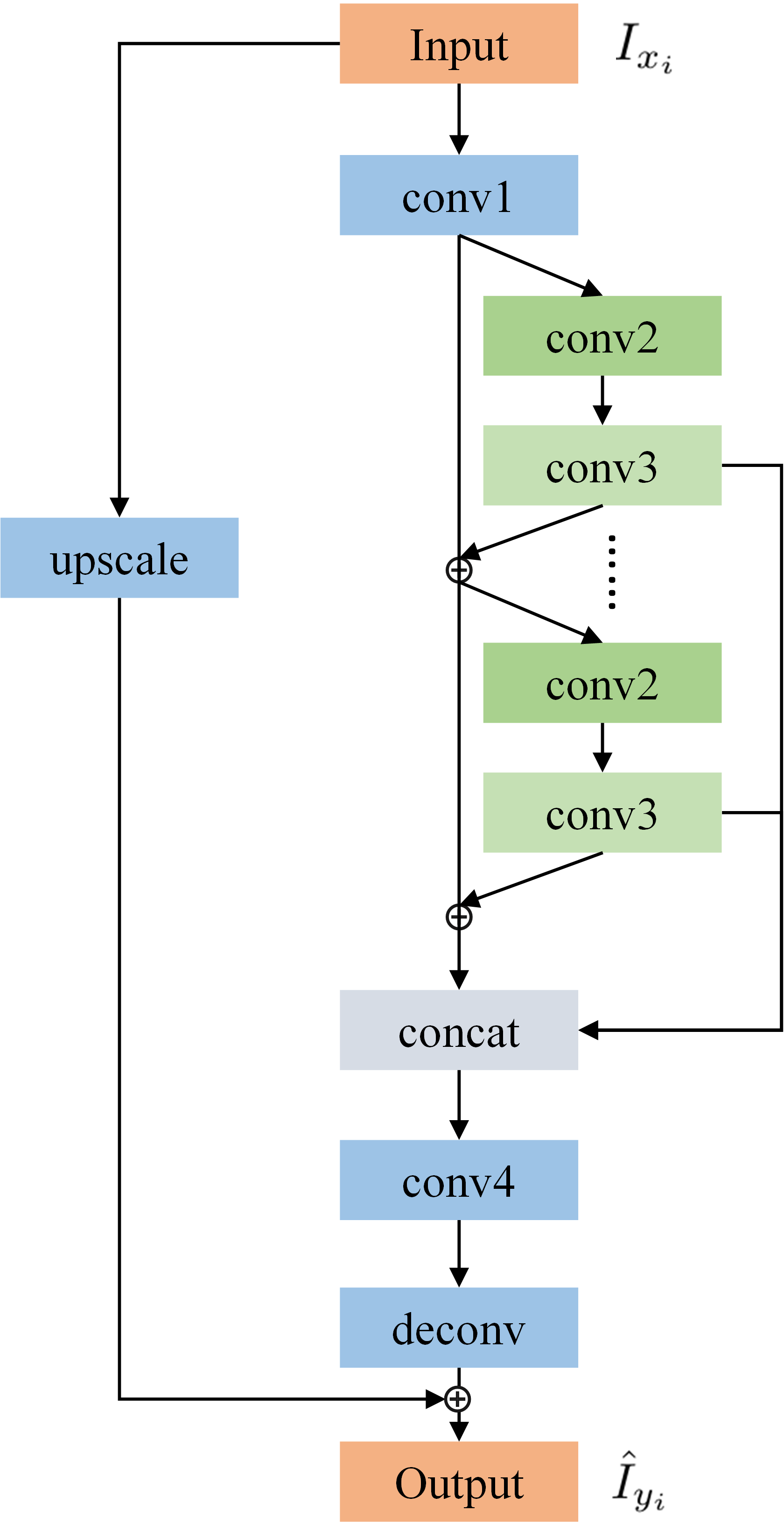}}\hskip 18pt
\subfloat[]{\label{fig:5c}
\includegraphics[width=0.24\textwidth]{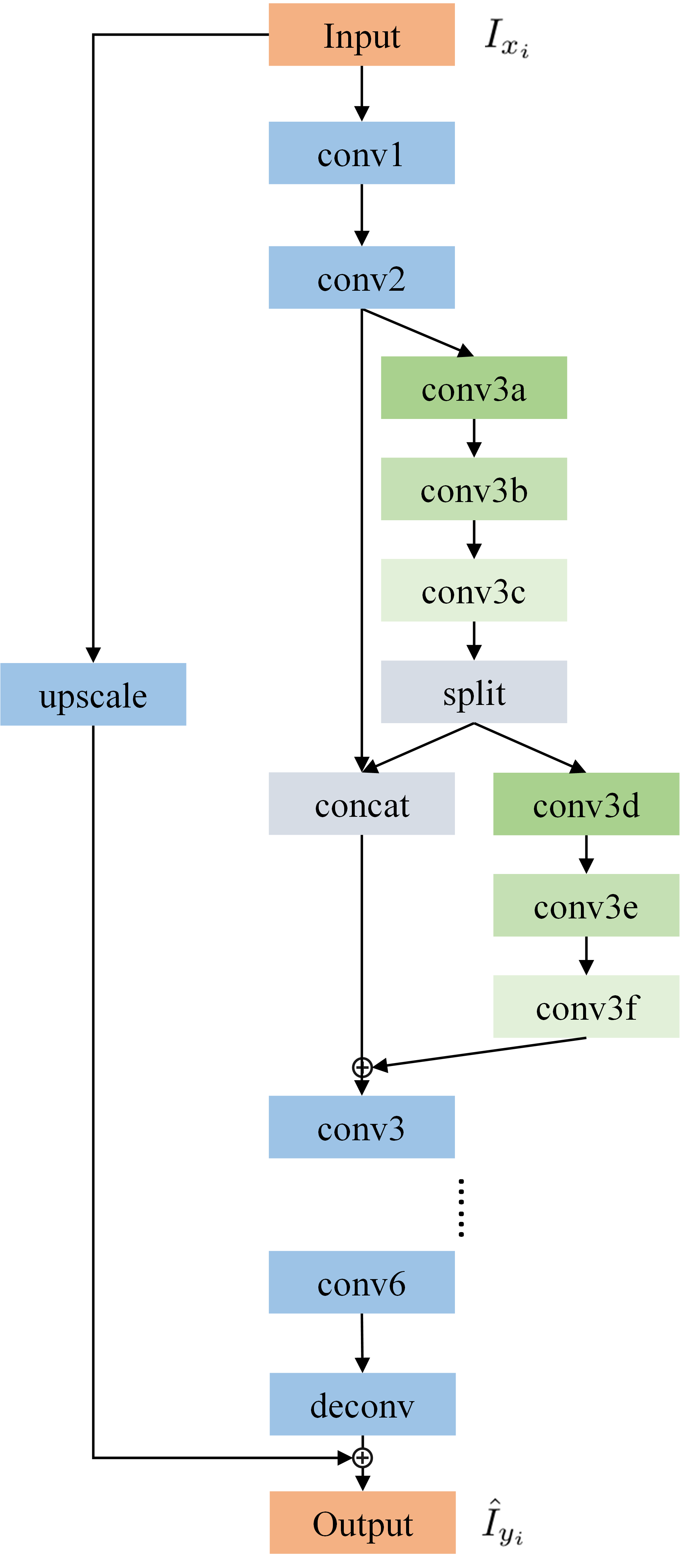}}
\caption{The schematic sketches for our (a) VDSR, (b) DRRN, and (c) IDN.}
\label{fig:5}
\end{center}
\vskip -0.15in
\end{figure*}

\begin{figure*}[h!]
\captionsetup[subfigure]{labelformat=empty}
\begin{center}
%

\subfloat[Input\hspace{3em} $u=2.6$]{
\includegraphics[width=0.07\textwidth]{images/tmp/Set5_ORI_2.6/butterfly_GT_ll.png}}\hskip 18pt
\subfloat[$u=0.0$, DRRN]{
\includegraphics[width=0.21\textwidth]{images/tmp/Set5_ORI_0/butterfly_GT_fk.png}}\hskip 18pt
\subfloat[$u=1.3$, DRRN]{
\includegraphics[width=0.21\textwidth]{images/tmp/Set5_ORI_1.3/butterfly_GT_fk.png}}\hskip 18pt
\subfloat[$u=2.6$, DRRN]{
\includegraphics[width=0.21\textwidth]{images/tmp/Set5_ORI_2.6/butterfly_GT_fk.png}}\hskip 18pt \\

\hskip 54pt
\subfloat[$u=0.0$, DRRN (DL-Net)]{
\includegraphics[width=0.21\textwidth]{images/tmp/Set5_DLNet_0/butterfly_GT_fk.png}}\hskip 18pt
\subfloat[$u=1.3$, DRRN (DL-Net)]{
\includegraphics[width=0.21\textwidth]{images/tmp/Set5_DLNET_1.3/butterfly_GT_fk.png}}\hskip 18pt
\subfloat[$u=2.6$, DRRN (DL-Net)]{
\includegraphics[width=0.21\textwidth]{images/tmp/Set5_DLNET_2.6/butterfly_GT_fk.png}}\hskip 18pt 

\caption{SISR results with the original DRRN and the retrained model powered by our method.} 
\label{fig:6}
\end{center}
\vskip -0.15in
\end{figure*}

\section{Qualitative Results for SISR}\label{sec:qualitative}
We also provide some qualitative results for SISR under multiple degradation settings. See Figure~\ref{fig:6} for more details.
We illustrate the luminance channel only to enable comparison between direct outputs of different network models.
It can be seen that our model show perceptually similar results under different degradation settings while the performance of original DRRN diminish significantly under $u=1.3$ and $u=2.6$.


\bibliographystyle{IEEEtran}
\bibliography{ref}

\begin{IEEEbiography}[{\includegraphics[width=1in,height=1.25in,clip,keepaspectratio]{images/yiwen.png}}]{Yiwen Guo} received the B.E. degree from Wuhan University, Wuhan, China, in 2011, and the Ph.D. degree from Tsinghua University, Beijing, China in 2016. He is a research scientist at Bytedance AI Lab, Beijing. Prior to this, he was a staff research scientist at Intel Labs China. His current research interests include computer vision, pattern recognition, and machine learning.
\end{IEEEbiography}

\begin{IEEEbiography}[{\includegraphics[width=1in,height=1.25in,clip,keepaspectratio]{images/luming.jpg}}]{Ming Lu} Ming Lu received the PhD degree in Information and Communication Engineering from Tsinghua University, Beijing, China, in 2019. He is currently a researcher at Intel Labs China. His research interests include computer vision and computer graphics. He is particularly interested in classification/detection, 3D face/body, and image restoration/synthesis.\end{IEEEbiography}

\begin{IEEEbiography}[{\includegraphics[width=1in,height=1.25in,clip,keepaspectratio]{images/WMZ.jpg}}]{Wangmeng Zuo} (M'09, SM'14) received the Ph.D. degree in computer application technology from the Harbin Institute of Technology, China, in 2007. From 2004 to 2006, he was a Research Assistant with the Department of Computing, The Hong Kong Polytechnic University. From 2009 to 2010, he was a Visiting Professor with Microsoft Research Asia. He is currently a Professor with the School of Computer Science and Technology, Harbin Institute of Technology. He has published over 90 papers in top-tier academic journals and conferences. His current research interests include image enhancement and restoration, image generation and editing, visual tracking, object detection, and image classification. He has served as a Tutorial Organizer in ECCV 2016, an Associate Editor of the IET Biometrics, and the Guest Editor of Neurocomputing, Pattern Recognition, IEEE Transactions on Circuits and Systems for Video Technology, and IEEE Transactions on Neural Networks and Learning Systems.
\end{IEEEbiography}

\begin{IEEEbiography}[{\includegraphics[width=1in,height=1.25in,clip,keepaspectratio]{images/zcs.jpeg}}]{Changshui Zhang} received the B.E. degree in mathematics from Peking University, Beijing, China, in 1986, and the M.S. and Ph.D. degrees in control science and engineering from Tsinghua University, Beijing, in 1989 and 1992, respectively. In 1992, he joined the Department of Automation, Tsinghua University, where he is currently a professor. His research interests include pattern recognition and machine learning. He is a Fellow member of the IEEE.
\end{IEEEbiography}

\begin{IEEEbiography}[{\includegraphics[width=1in,height=1.25in,clip,keepaspectratio]{images/yurong.jpg}}]{Yurong Chen} received the B.S. and Ph.D. degrees from Tsinghua University, Beijing, China, in 1998 and 2002, respectively. He joined Intel in 2004 after completing the postdoctoral research in the Institute of Software, CAS, where he is currently a Principal Research Scientist and Director of Cognitive Computing Lab at Intel Labs China, responsible for leading visual cognition and machine learning research for Intel platforms. He received one “Intel China Award” and 3 Intel Labs Academic Awards – “Gordy Awards” for delivering leading visual analytics and understanding technologies to impact Intel platforms/solutions. He has published over 60 papers and holds over 50 issued/pending patents. 
\end{IEEEbiography}